\pdfoutput=1
\documentclass[lettersize,journal]{IEEEtran}

\usepackage{amsmath,bm,amssymb}
\usepackage{pifont}

\usepackage[table]{xcolor}

\usepackage{multirow}
\usepackage{booktabs}
\usepackage{orcidlink}
\usepackage{overpic}
\usepackage{hyperref}
\hypersetup{
     colorlinks=true,
     linkcolor=blue,
     citecolor=blue,
     filecolor=blue,
     urlcolor=magenta,
}
\usepackage[nocompress]{cite}

\usepackage{makecell}
\usepackage{siunitx}
\usepackage[linesnumbered,ruled,vlined]{algorithm2e}
\usepackage[capitalise]{cleveref}
\crefname{section}{Sect.}{Sects.}
\Crefname{section}{Sect.}{Sects.}
\crefname{figure}{Fig.}{Figs.}
\Crefname{figure}{Fig.}{Figs.}
\crefname{table}{Tab.}{Tabs.}
\Crefname{table}{Tab.}{Tabs.}
\crefname{appendix}{}{Appendixes}
\Crefname{appendix}{}{Appendixes}

\crefformat{equation}{Eq.~{#2(#1)#3}}
\newtheorem{theorem}{Theorem}
\newtheorem{proof}{Proof}

\usepackage{subcaption}
\usepackage{tikz}

\newcommand{\myPara}[1]{%
  \textbf{#1}%
}

\newcommand{\MSsubsection}[2]{%
\subsection{#2}\label{#1}%
}

\newcommand{\MSfigcaption}[3]{%
\caption{#2}
\label{#1}#3
}

\makeatletter
\newcommand{\MSphantomsubcaptions}[1]{%
  \begingroup
  \let\caption@@stepcounter\@gobble
  #1%
  \endgroup
}
\makeatother

\newcommand{\MStabcaption}[2]{\caption{#2}\label{#1}}

\usepackage{array}
\usepackage{arydshln}
\newlength{\myMheight}
\newcommand{\imgcirc}[4]{%
  \raisebox{-0.5\height}{%
    \begin{overpic}[width=0.185\linewidth,percent]{#1}
      \put(#2,#3){\tikz\draw[black,line width=1pt] (0,0) circle (#4pt);}
    \end{overpic}%
  }%
}
\newcommand{\imgplain}[1]{%
  \raisebox{-0.5\height}{\includegraphics[width=0.185\linewidth]{#1}}%
}

\begin{document}

\title{
  Inclusive Hand Pose Estimation
  via Topology-Aware Distillation and Progressive Kinematic Reasoning
}

\author{
  Kanglei Zhou$^{\orcidlink{0000-0002-4660-581X}}$,
  Zikai Hao$^{\orcidlink{0009-0004-5968-7764}}$,
  Liyuan~Wang$^{\orcidlink{0009-0002-7797-325X}}$,
  and
  Xiaohui Liang$^{\orcidlink{0000-0001-6351-2538}}$,~\IEEEmembership{Member,~IEEE}
  \thanks{
    \copyright~2026 IEEE. Personal use of this material is permitted. Permission from IEEE must be obtained for all other uses, in any current or future media, including reprinting/republishing this material for advertising or promotional purposes, creating new collective works, for resale or redistribution to servers or lists, or reuse of any copyrighted component of this work in other works.
    This work was supported by the National Natural Science Foundation of China under Project 62272019, the China Postdoctoral Science Foundation under Project 2025M781489, and the Postdoctoral Fellowship Program (Grade C) of China Postdoctoral Science Foundation under Project GZC20260911. 
    \textit{(Corresponding author: Xiaohui Liang.)}
  }%
  \IEEEcompsocitemizethanks{
    \IEEEcompsocthanksitem K. Zhou and L. Wang are with the Department of Psychological and Cognitive Sciences, Tsinghua University, Beijing, China.
    E-mail: zhoukanglei@tsinghua.edu.cn, liyuanwang@tsinghua.edu.cn
    \IEEEcompsocthanksitem Z. Hao is with the State Key Laboratory of Virtual Reality Technology and Systems, Beihang University, Beijing, China.
    E-mail: zy2306413@buaa.edu.cn
    \IEEEcompsocthanksitem X. Liang is with the State Key Laboratory of Virtual Reality Technology and Systems, Beihang University, Beijing, China, and also with Zhongguancun Laboratory, Beijing, China.
    E-mail: liang\_xiaohui@buaa.edu.cn
  }%
}

\markboth{IEEE Transactions on Visualization and Computer Graphics}{Zhou \MakeLowercase{\textit{et al.}}: Inclusive Hand Pose Estimation via Topology-Aware Distillation and Progressive Kinematic Reasoning}

\maketitle

\begin{abstract}%
  Hand pose estimation aims to recover the spatial configuration of hand joints from visual observations and serves as a fundamental perception module for downstream systems that rely on complete and topologically consistent hand representations.
  Most existing methods are developed under the assumption of fully articulated hand structures and tend to produce unstable or implausible predictions when parts of the kinematic structure are persistently absent, resulting in degraded performance or even system failure despite the availability of partial observations.
  To systematically investigate this challenge, we construct \textbf{IncNYU}, a physiologically constrained benchmark that simulates structural incompleteness through morphological modification of complete hand models and retains reference locations for both intact and synthetically removed regions.
  Based on this benchmark, we propose \textbf{InsHaPE}, an \underline{In}clu\underline{s}ive \underline{Ha}nd \underline{P}ose \underline{E}stimation framework designed to remain robust under structural incompleteness while preserving performance on standard, fully articulated hands.
  InsHaPE is built upon a hierarchical kinematic modeling paradigm and introduces two complementary components.
  First, a topology-aware distillation strategy transfers joint-level and relational priors from fully articulated hand representations, providing anatomically consistent initialization under missing structures.
  Second, a progressive kinematic reasoning module performs structured inference along the kinematic chain, enabling complete and anatomically consistent pose estimation under severe structural variations.
Extensive experiments on NYU/IncNYU, together with external validation on MSRA, demonstrate that InsHaPE improves robustness under structural incompleteness while maintaining competitive performance on complete hands.
  Additional system-level evaluations further show that InsHaPE improves robustness for downstream applications that require fixed-topology hand pose representations.
  Our code is available at \url{https://github.com/tomhao555/InsHaPE}.
\end{abstract}


\section{Introduction}

\IEEEPARstart{H}{and} pose estimation~\cite{ohkawa2023efficient,wan2018dense,moon2022accurate,salter2024emg2pose} aims to recover the 2D/3D spatial configuration of hand joints from sensor data and serves as a core perception module in many Mixed Reality (MR) systems, supporting applications such as gesture-based interaction, virtual object manipulation, and immersive user interfaces~\cite{zhao2024gaussianhand,shi2022grasping}. In these systems, hand pose estimation acts as an intermediate representation that bridges low-level visual sensing and high-level interaction logic \cite{zhou2026comprehensive}. Consequently, accurate and stable joint predictions are essential for downstream components that rely on complete and topologically consistent hand representations~\cite{zhou2021stgae}.

\begin{figure}
  \includegraphics[width=\linewidth,clip,trim=120 45 120 100]{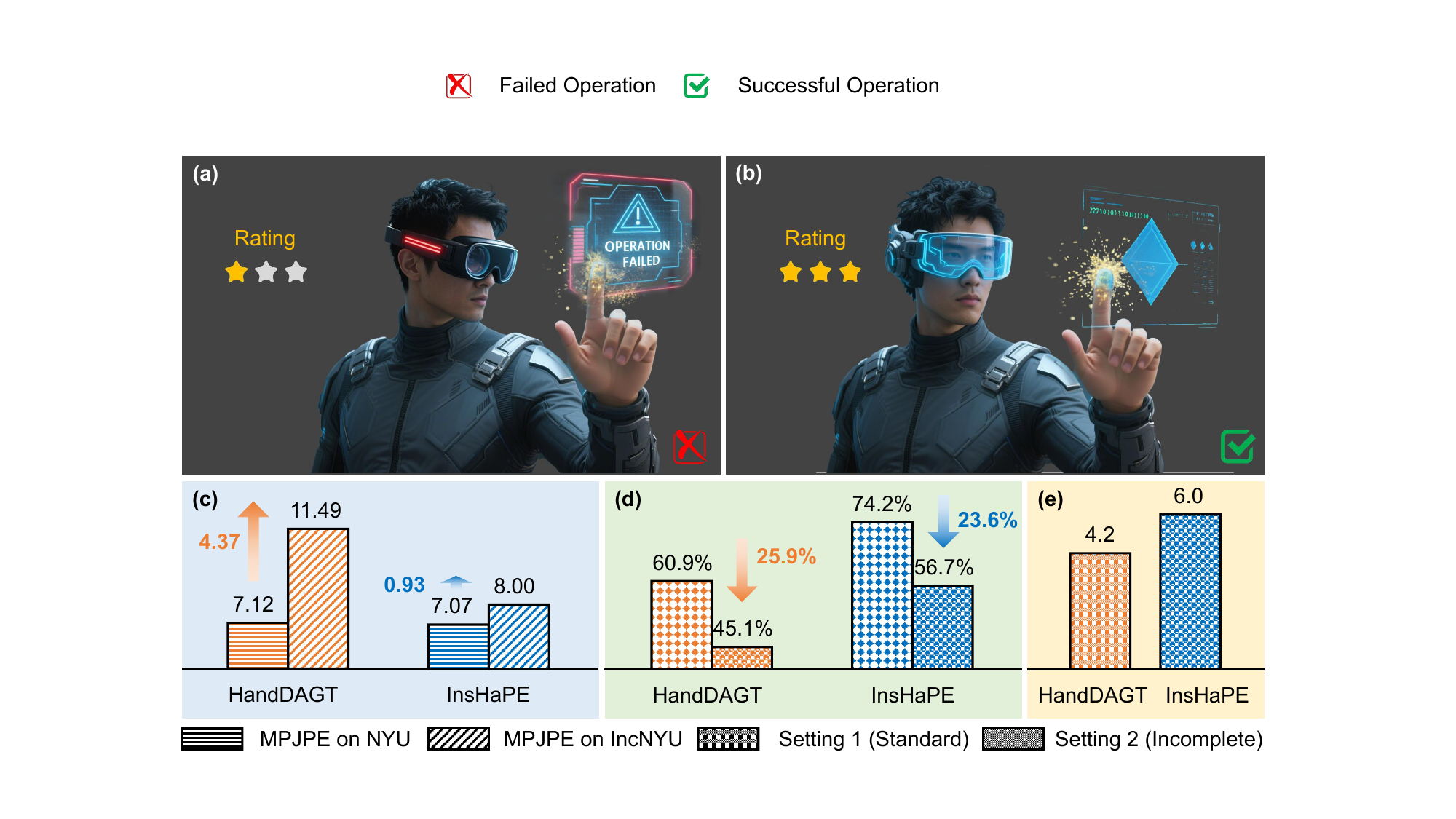}%
  \caption{
    System-level comparison between a representative baseline method~\cite{cheng2024handdagt} and the proposed InsHaPE framework under standard and structurally incomplete settings.
    (a) The baseline method produces erroneous or implausible joint predictions when parts of the hand structure are missing.
    (b) In contrast, InsHaPE maintains topologically consistent and complete pose estimation under the same conditions.
    (c) Quantitative comparison of pose estimation accuracy (MPJPE) on NYU and IncNYU, where the baseline exhibits a substantial performance degradation (from 7.12 to 11.49 MPJPE), while InsHaPE shows only a marginal increase (from 7.07 to 8.00).
    (d) Gesture recognition accuracy under complete-hand and simulated-incomplete observation conditions, where InsHaPE outperforms the baseline in both settings (74.2\% vs.\ 60.9\%, and 56.7\% vs.\ 45.1\%) and the arrows indicate the relative accuracy drop from the complete to the incomplete setting.
    (e) Subjective usability scores reported by participants in a controlled system-level evaluation, reflecting improved reliability and compatibility of InsHaPE with downstream applications that require fixed-topology hand pose representations.
  }
  \label{fig:teaser}{%
    \MSphantomsubcaptions{%
      \phantomsubcaption\label{fig:teaser-a}%
      \phantomsubcaption\label{fig:teaser-b}%
      \phantomsubcaption\label{fig:teaser-c}%
      \phantomsubcaption\label{fig:teaser-d}%
      \phantomsubcaption\label{fig:teaser-e}%
    }%
  }%
\end{figure}

Despite recent advances, most existing hand pose estimation methods~\cite{ren2023two,cheng2024handdiff,cheng2024handdagt,jiang2023a2j} are developed under the implicit assumption of fully articulated hand structures.
This assumption is violated by permanent anatomical absence; persistent sensing failures can eliminate reliable evidence for particular joints.
Under such structural incompleteness, existing methods tend to produce unstable or implausible predictions, as evidenced by the substantial error increase in \cref{fig:teaser-c}.
These errors further propagate to downstream MR systems and third-party applications that rely on fixed-topology pose representations, leading to degraded performance~\cite{mo2021gesture,zhao2022place,zhou2023mixed,ohkawa2023assemblyhands}, as reflected by the drop in gesture recognition accuracy in \cref{fig:teaser-d}.
More importantly, since standard interaction APIs (e.g., OpenXR) assume complete and consistent hand pose inputs, unreliable estimation under structural incompleteness directly compromises compatibility with existing pipelines.
Addressing structural incompleteness at the pose estimation stage therefore supports robust downstream interaction without requiring changes to fixed-topology application interfaces.

\begin{figure}
  \centering
  \includegraphics[width=\linewidth,clip,trim=126 140 126 140]{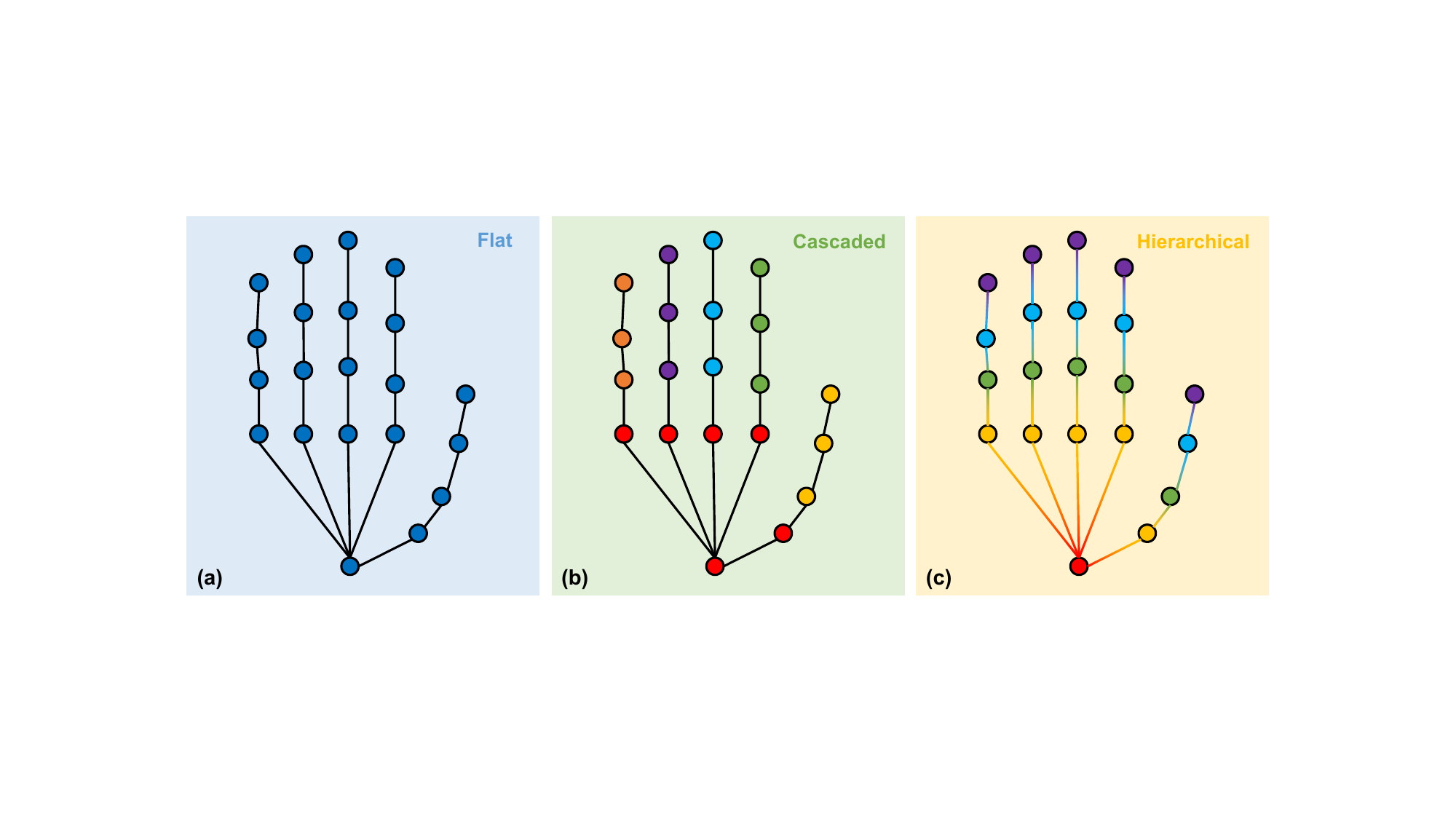}
  \caption{Comparison of three hand pose modeling paradigms.
    (a) Flat joint-wise modeling, where all joints share a single parameterization and are regressed independently.
    (b) Cascaded (finger-wise) modeling, which performs part-level cascaded prediction across the palm and fingers but lacks explicit intra-finger kinematic hierarchy.
    (c) Our hierarchical kinematic modeling, which conducts progressive root-to-tip inference and explicitly encodes kinematic dependencies, enabling robust pose estimation under structural incompleteness.
  }
  \label{fig:method-cmp}
  \MSphantomsubcaptions{%
    \phantomsubcaption\label{fig:method-cmp-a}
    \phantomsubcaption\label{fig:method-cmp-b}
    \phantomsubcaption\label{fig:method-cmp-c}
  }
\end{figure}

These failures are not incidental but stem from fundamental limitations in existing data, modeling assumptions, and reasoning strategies.
First, most existing benchmarks~\cite{li2023renderih} are collected from fully articulated hands and implicitly enforce complete kinematic supervision, which biases trained models toward predicting full joint sets and leaves them unexposed to missing-structure cases during learning.
Second, although several approaches~\cite{zhou2024multi,ye2018occlusion,meng20223d,liu2022explicit,park2022handoccnet} have been proposed to address occlusion, they generally treat missing observations as transient or noisy and rely on temporal smoothing, spatial context, or feature hallucination.
Such assumptions do not hold under persistent structural absence, where kinematic elements are permanently removed and no valid observation or temporal cue exists to support reliable recovery.
Third, although existing methods~\cite{sun2015cascaded,deng2022recurrent,cheng2024handdagt,leng2021stable} have explored cascaded refinement or topology-aware feature modeling, they usually exploit hand topology at the feature aggregation stage rather than explicitly constraining the final prediction process along the anatomical kinematic chain, as illustrated in \cref{fig:method-cmp-a,fig:method-cmp-b}.
Under structural incompleteness, weak distal-joint evidence makes joint prediction and unconstrained refinement unstable, motivating prediction-level kinematic constraints.

To systematically study hand pose estimation under structural incompleteness, we consider scenarios in which parts of the hand kinematic structure are persistently absent, a setting that is largely overlooked in existing benchmarks.
To this end, we construct a novel benchmark by extending the widely used NYU dataset~\cite{tompson2014real}.
Using physiologically constrained morphological processing, we generate diverse forms of structural absence, including missing fingers and missing joint segments, while preserving realistic hand geometry and articulation.
The resulting benchmark, termed \textbf{IncNYU}, serves as a controlled and repeatable stress test for evaluating the robustness and generalization of hand pose estimation methods under incomplete kinematic structures.

To address these challenges, we propose \textbf{InsHaPE}, an \underline{In}clu\underline{s}ive \underline{Ha}nd \underline{P}ose \underline{E}stimation framework designed to remain robust under structural incompleteness while preserving performance on standard, fully articulated hands.
Here, ``inclusive'' denotes the design goal of accommodating diverse hand kinematic structures while maintaining consistent pose representations required by downstream MR applications.
InsHaPE adopts a hierarchical kinematic modeling paradigm (see \cref{fig:method-cmp-c}) that captures dependencies along the articulated hand structure.
Based on this formulation, InsHaPE consists of two complementary components.
First, a \textbf{topology-aware distillation} strategy transfers structural priors learned from fully articulated hand data to provide a reliable and anatomically consistent initialization under missing structures.
Building on this representation, a \textbf{progressive kinematic reasoning} module infers joints along the kinematic chain, enabling complete and anatomically consistent pose estimation even when multiple kinematic elements are absent.

Extensive experiments show that InsHaPE substantially improves performance on IncNYU while maintaining competitive accuracy on NYU, with additional validation on MSRA further supporting its generalizability.
In addition, a prototype MR system and user study demonstrate the effectiveness of InsHaPE for reliable downstream MR interaction under structural incompleteness.
Our contributions are threefold:

\begin{itemize}
  \item We formulate hand pose estimation under structural incompleteness and introduce \textbf{IncNYU}, a physiologically constrained benchmark for systematic evaluation.
  \item We propose a \textbf{topology-aware distillation} strategy that effectively transfers structural priors and improves generalization under incomplete kinematic structures.
  \item We introduce a \textbf{progressive kinematic reasoning} module that hierarchically infers hand joints to enable robust and topologically consistent pose estimation.
\end{itemize}

The remainder of this paper is organized as follows.
\cref{sec:related_work} reviews related work on hand pose estimation.
\cref{sec:problem} formulates hand pose estimation under structural incompleteness and introduces the IncNYU benchmark for systematic evaluation.
\cref{sec:method} presents the proposed InsHaPE framework.
\cref{sec:theory} provides a supporting theoretical analysis that explains why the proposed method remains effective under both structurally incomplete and standard settings.
\cref{sec:exp} presents extensive experimental results.
\cref{sec:system} reports system-level evaluations and a user study.
Finally, \cref{sec:conclusion} concludes the paper.

\section{Related Work} \label{sec:related_work}
This section reviews prior work on hand pose estimation, including occlusion-aware methods and structure-aware modeling strategies, and discusses existing efforts toward inclusive interaction in MR that motivate our problem setting.

\subsection{Hand Pose Estimation}
Hand pose estimation methods can be broadly categorized according to input modality and modeling strategy.
Early approaches primarily relied on 2D convolutional neural networks applied to depth images~\cite{du2019crossinfonet,fang2020jgr,ge2016robust,ren2021pose,tompson2014real}.
While computationally efficient, these methods struggled to capture the full 3D articulated structure of the hand and were sensitive to viewpoint variations.
Subsequent 3D CNN-based approaches~\cite{ge20173d,moon2018v2v} addressed this limitation by operating on voxelized depth representations, enabling volumetric reasoning at the cost of increased memory consumption and computational complexity.
To better exploit geometric information, point cloud-based methods, such as PointNet~\cite{qi2017pointnet} and HandPointNet~\cite{ge2018hand}, directly process 3D point sets to estimate hand pose.
Although effective in capturing spatial structure, these methods often rely on sparse sampling or simplified representations to maintain computational efficiency, which may limit estimation accuracy for fine-grained joint configurations.
More recently, multimodal approaches~\cite{lin2023cross,ge2025coarse} that integrate 2D image cues with 3D point cloud features have been proposed to leverage complementary appearance and geometric information.
In parallel, generative modeling techniques, particularly diffusion models~\cite{ho2020denoising,sohl2015deep,ivashechkin2023denoising}, have been explored for hand pose estimation to better model uncertainty in the 2D-to-3D lifting process.
Most existing diffusion-based methods adopt a two-stage pipeline, first estimating 2D keypoints and then refining 3D pose distributions, while more recent single-stage formulations directly condition on 3D representations for improved efficiency.
Despite substantial progress, existing methods and benchmarks are predominantly developed under the assumption of fully articulated and anatomically intact hands.
Consequently, their robustness and generalization to scenarios involving persistent structural variations, such as missing joints or abnormal hand morphology, remain limited, motivating the need for structure-aware modeling.

\subsection{Occlusion-Aware Hand Pose Estimation}

Robust hand pose estimation under occlusion has long been a central challenge, motivating a wide range of occlusion-aware modeling strategies.
Early work, such as Hierarchical Mixture Density Networks (HMDN)~\cite{ye2018occlusion}, introduced probabilistic formulations to model uncertainty in self-occluded joints from depth images, enabling multimodal predictions when visual evidence is incomplete.
Subsequent approaches incorporated multiview or multimodal cues to enhance robustness, for example by combining amodal segmentation and hand de-occlusion modules to restore full hand skeletons under severe occlusion~\cite{meng20223d}.
More recent methods, including HandOccNet~\cite{park2022handoccnet}, employ Transformer-based architectures to inject contextual information from visible regions into occluded areas, significantly improving 3D hand mesh recovery.
Beyond purely visual cues, cross-modal approaches have also been explored.
For instance, OCHID-Fi~\cite{zhang2023ochid} integrates RF sensing with vision to estimate hand pose even when parts of the hand fall outside the camera's field of view, highlighting the potential of complementary sensing modalities.
In parallel, masked modeling and attention-based frameworks, such as OccRobNet~\cite{garg2025occrobnet}, HOMAE~\cite{yang2025occlusion}, and HOCAT~\cite{cai2025geometry}, leverage self-attention, cross-attention, and explicit feature decoupling to infer occluded joints through context-aware reasoning.
Comprehensive surveys~\cite{ohkawa2023efficient,woo2023survey} further summarize advances in occlusion-aware annotation strategies, learning paradigms, and dataset design.
Despite notable progress, a fundamental assumption underlies most occlusion-aware methods: all hand joints are physically present but temporarily hidden from observation.
Occlusion is therefore treated as a transient or recoverable condition, where missing information can be inferred from spatial context, temporal continuity, or cross-modal cues.
In contrast, scenarios involving structural incompleteness, such as anatomically missing or permanently deformed joints, violate this assumption and provide no valid visual or temporal evidence for recovery.
As a result, occlusion-aware techniques are not directly applicable to true structural absence, highlighting a critical gap between occlusion handling and pose estimation under incomplete kinematic structures.

\subsection{Inclusive Design and Accessibility in MR}
Recent years have seen growing interest in improving accessibility and inclusivity in MR systems.
Prior work has examined inclusive design principles and system-level frameworks, such as ``Inclusive Immersion,'' which synthesize academic research and industry practices to identify key accessibility challenges in immersive environments~\cite{dudley2023inclusive}.
Complementary surveys further analyze barriers faced by users with diverse impairments, including visual, auditory, and motor limitations, and categorize software, hardware, and ethical challenges through collaboration with domain experts and people with disabilities~\cite{inclusiveAgenda2023,creed2024inclusive}.
On the technical side, existing studies have explored device-level inclusivity, adaptive system architectures, and user-centered interaction design to improve usability across heterogeneous user populations~\cite{biswas2021adaptive,krug2024don,kim2025immersive}.
Additional efforts have investigated accessibility in application-specific domains such as healthcare and large-scale virtual environments, proposing high-level frameworks that integrate assistive technologies, policy considerations, and interoperable standards~\cite{othman2024accessible,radanliev2024accessibility}.
Despite these advances, most prior work in inclusive MR focuses on interaction design, interface adaptation, or system-level accessibility, while relatively little attention has been paid to the robustness of low-level perception modules that underpin many MR applications.
In particular, hand pose estimation, which serves as a critical interface between user motion and system interaction, is typically assumed to operate on fully articulated hands.
This gap highlights the need for perception methods that remain reliable under diverse hand kinematic structures.

\section{Problem Formulation and Benchmark} \label{sec:problem}
This section formulates hand pose estimation under structural incompleteness and presents the IncNYU benchmark.

\subsection{Problem Formulation}
We study hand pose estimation under \emph{structural incompleteness}, where parts of the hand kinematic structure are anatomically absent rather than temporarily occluded. We additionally use persistent sensor-induced missing observations as a controlled system-level stress test, while distinguishing this condition from anatomical incompleteness. In both settings, the missing evidence cannot be recovered from the current visual input or short-term temporal cues.
Given an input depth image $\mathbf{I} \in \mathbb{R}^{H \times W}$, the goal is to estimate the 3D locations of a predefined set of $K$ hand joints,
\begin{equation}
  \hat{\mathcal{P}} = \{\hat{\bm{p}}_1, \hat{\bm{p}}_2, \ldots, \hat{\bm{p}}_K\}, \quad \hat{\bm{p}}_k \in \mathbb{R}^3,
\end{equation}
where $\hat{\mathcal{P}}$ has the fixed topology required by MR systems.

Unlike conventional hand pose estimation~\cite{cheng2024handdagt,cheng2024handdiff,cheng2023handr2n2}, we consider cases in which a subset of joints is anatomically absent because of permanent structural variations (e.g., finger amputation or partial joint loss).
We denote the intact joints by $\mathcal{K}_{\mathrm{intact}} \subseteq \{1,\ldots,K\}$ and the structurally absent joints by $\mathcal{K}_{\mathrm{absent}} = \{1,\ldots,K\} \setminus \mathcal{K}_{\mathrm{intact}}$.
For each $k \in \mathcal{K}_{\mathrm{absent}}$, neither a valid visual observation nor a corresponding physical articulation exists in $\mathbf{I}$.
Nevertheless, the estimator must output a complete pose over all $K$ joints, as many downstream MR applications~\cite{zhou2023mixed,zhao2024gaussianhand,zhou2021stgae} rely on fixed-topology pose representations and cannot accommodate variable joint sets.
Accordingly, our task is to infer a \emph{topologically complete and anatomically coherent} pose representation that remains compatible with existing interaction pipelines without requiring modifications to their fixed-topology interfaces.

Formally, we learn the following mapping:
\begin{equation}
  f_\theta : \mathbf{I} \rightarrow \hat{\mathcal{P}},
\end{equation}
that is defined over both intact and structurally absent joint indices.
Predictions should align with observed geometry for intact joints and provide kinematically plausible placeholders for the structurally absent hand regions.

This setting differs from occlusion-aware hand pose estimation, where joints are assumed to exist but be temporarily hidden.
In contrast, structural incompleteness corresponds to permanent removal of kinematic elements, for which no direct observation or temporal recovery is available.
We next introduce the physiologically constrained IncNYU benchmark.

\subsection{The Benchmark of IncNYU}

There is currently no publicly available benchmark that provides paired depth observations and joint annotations for hands with finger amputations, and the rarity of real-world cases makes large-scale data collection challenging.
A straightforward alternative is to synthesize amputation-like samples by adapting occlusion-oriented augmentation strategies.
However, common occlusion augmentations~\cite{choi2022learning,everingham2011pascal,rezaei2023trihorn} typically overlay random masks and do not respect anatomical constraints.
In contrast, physiologically valid amputations follow structural rules, e.g., an intermediate phalanx can only be absent if the distal phalanx is also missing.
Ignoring such constraints would lead to unrealistic samples.

To address this limitation, we construct a physiologically constrained synthetic amputation benchmark in the 3D domain based on the NYU dataset~\cite{tompson2014real}.
We focus on finger amputations, which are commonly reported in hand injuries~\cite{kawaiah2020fingertip}.
When generating an amputated sample, each finger is selected with probability $1/2$. Of the selected finger amputations, $2/3$ are assigned to fingertip loss, while the remainder are distributed randomly across other segments.

\begin{figure}
  \centering
  \includegraphics[width=\linewidth,clip,trim=120 185 120 160]{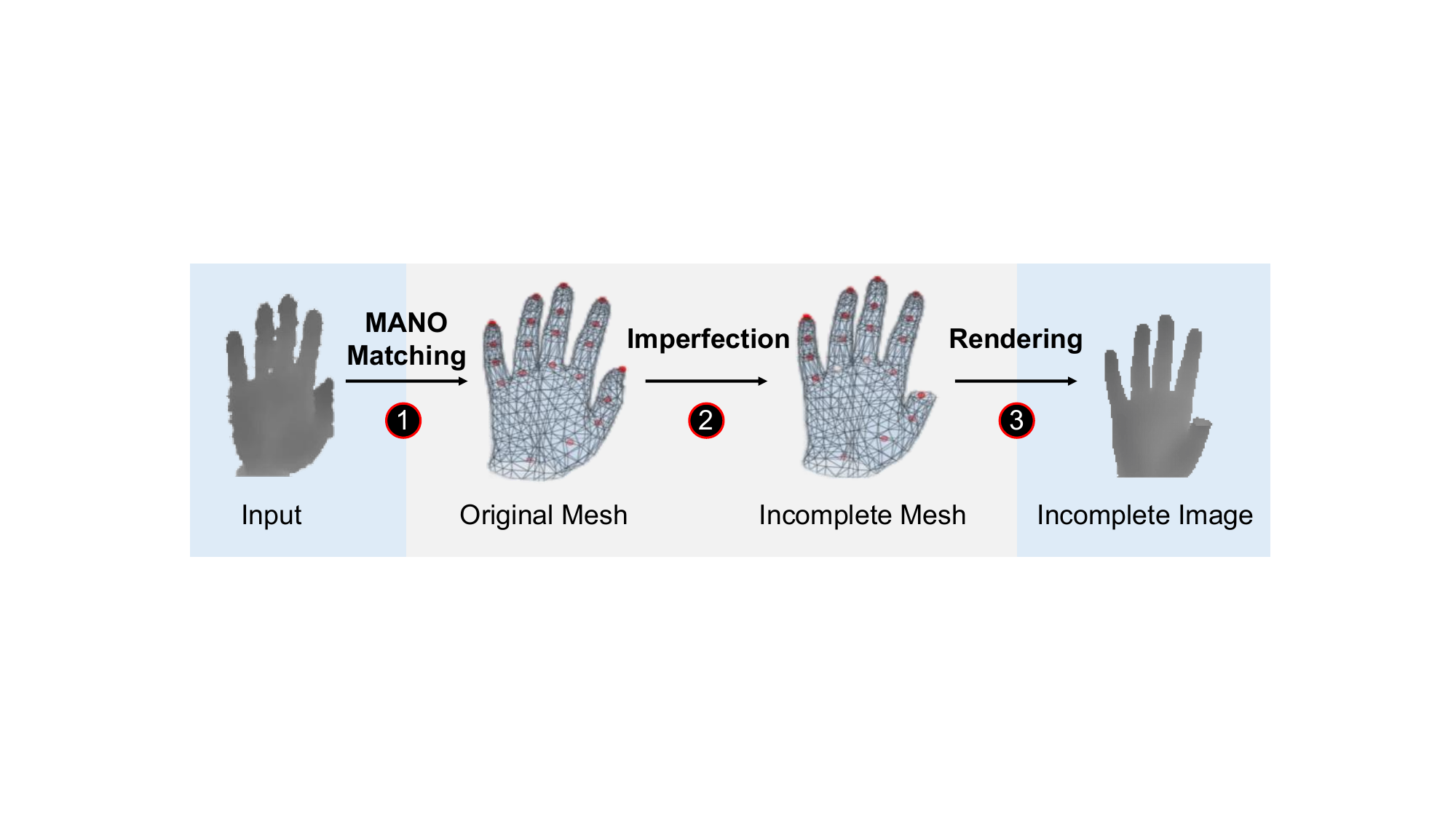}
  \caption{IncNYU construction pipeline.
    Unlike mask-based occlusion augmentation, we simulate finger amputations in 3D by removing MANO mesh segments according to anatomical rules and render the resulting incomplete hands as depth images.}
  \label{fig:data}
\end{figure}

The construction pipeline is illustrated in \cref{fig:data}.
First, we fit MANO meshes~\cite{armagan2020measuring} to NYU samples and manually annotate finger regions to establish a mapping between region identifiers and mesh vertices.
For each sample, amputation regions are selected according to the above probabilities.
We then remove the corresponding mesh vertices and update mesh faces, recursively deleting from distal to proximal segments to ensure physiological consistency, resulting in a 3D amputated hand mesh.
Finally, we render the modified meshes using the original NYU camera parameters. The resulting depth images constitute the proposed \textbf{IncNYU} benchmark.

For IncNYU, the training set retains 10\% of the original NYU dataset (7,275 depth maps), with 20\% amputated and 80\% intact hands.
The test set contains 8,252 amputated depth maps, matching the NYU test set in size.
Among the joints in the 14-keypoint evaluation subset, 21\% are labeled as structurally absent.
Following previous work~\cite{cheng2024handdagt}, the model predicts 23 keypoints; all NYU and IncNYU metrics use the standard 14-keypoint subset.
This design preserves realistic geometry and enables evaluation under structural incompleteness.

\section{InsHaPE: Inclusive Hand Pose Estimation} \label{sec:method}
This section presents the proposed InsHaPE framework.
We first introduce the overall architecture and design motivation, followed by detailed descriptions of its two core components: Topology-Aware Distillation and Progressive Kinematic Reasoning.
Finally, we describe the complete training procedure.

\begin{figure*}
  \centering
  \includegraphics[width=\linewidth,clip,trim=0 110 0 90]{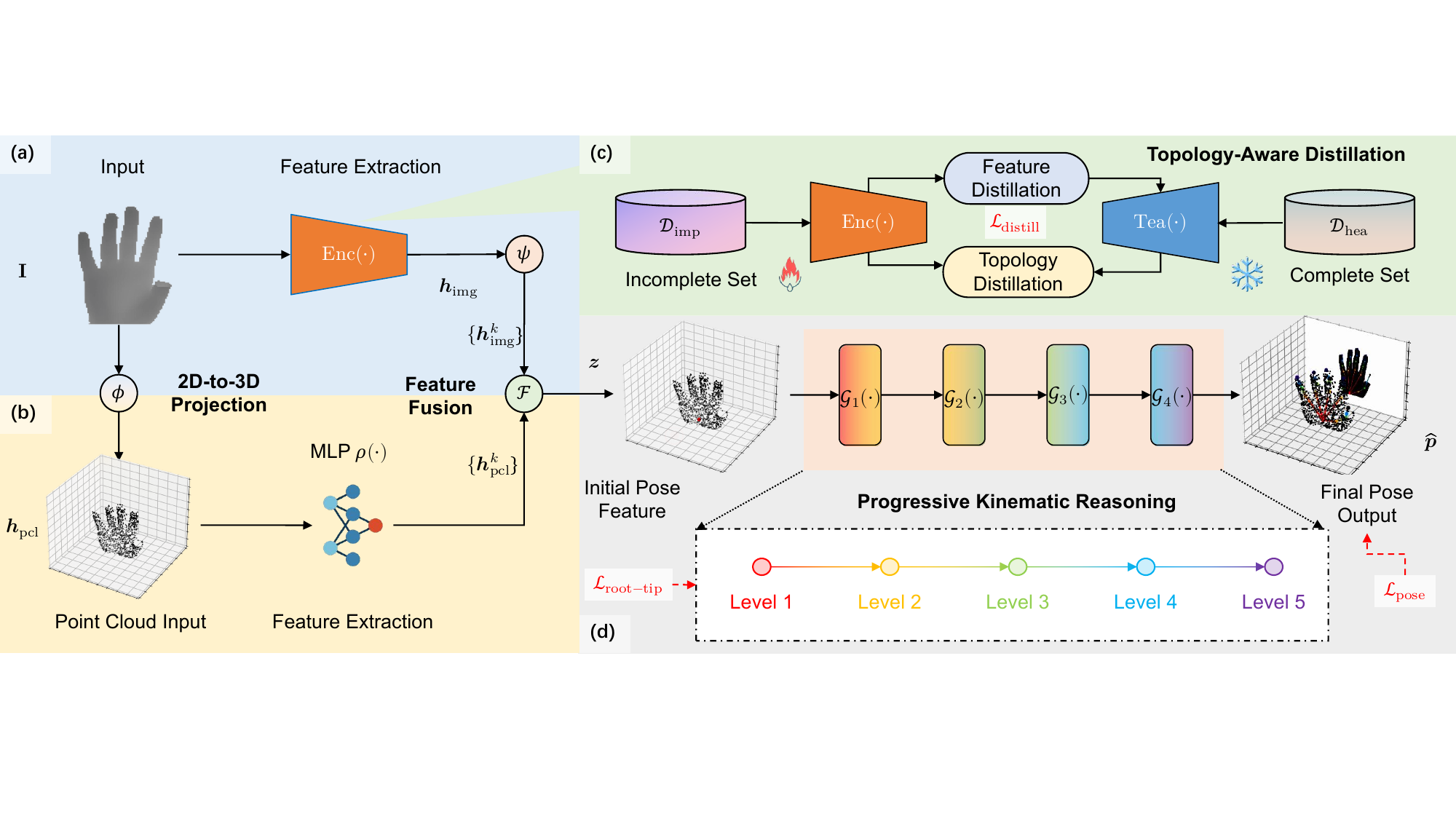}%
  \caption{Framework overview.
    \subref{fig:framework-a} Appearance features are extracted from the input depth image.
    \subref{fig:framework-b} Geometric point-cloud features are obtained from valid depth pixels through 2D-to-3D back-projection.
    \subref{fig:framework-c} Topology-aware distillation transfers structural priors from complete hand structures.
    \subref{fig:framework-d} Progressive kinematic reasoning performs hierarchical root-to-tip inference for complete pose estimation.
    \includegraphics[height=\myMheight]{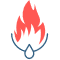} and \includegraphics[height=\myMheight]{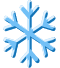} denote trainable and frozen backbones during topology-aware distillation, respectively.
  }
  \label{fig:framework}{%
    \MSphantomsubcaptions{%
      \phantomsubcaption\label{fig:framework-a}%
      \phantomsubcaption\label{fig:framework-b}%
      \phantomsubcaption\label{fig:framework-c}%
      \phantomsubcaption\label{fig:framework-d}%
    }%
  }
\end{figure*}

\subsection{Motivation and Framework}

\myPara{Motivation.}
Most existing hand pose estimation methods~\cite{ren2023two,cheng2024handdiff,cheng2024handdagt} are developed under the assumption of \emph{complete hand structures} and therefore struggle when parts of the kinematic structure are persistently absent.
As a result, MR applications that rely on fixed-topology hand pose representations often suffer from unstable interaction and reduced accessibility.
To support inclusive interaction, it is essential to estimate a complete and anatomically plausible set of hand keypoints even under structural incompleteness.
Motivated by these challenges, we propose \textbf{InsHaPE}, an inclusive hand pose estimation framework that is explicitly designed to handle structural incompleteness while remaining compatible with fixed-topology hand pose representations.

\myPara{Core Idea.}
The core idea of InsHaPE is to first distill topology-aware priors from complete hand structures and then retain these priors while learning hierarchical kinematic reasoning for structurally incomplete inputs.
By explicitly encoding kinematic hierarchy during inference, this design departs from joint-wise or weakly structured regression and enables robust, anatomically consistent pose estimation even when parts of the hand structure are permanently absent.

\myPara{Framework Overview.}
As illustrated in \cref{fig:framework}, InsHaPE is a unified framework for robust and inclusive hand pose estimation under structural incompleteness.
The framework is designed to jointly exploit visual appearance and explicit structure, while explicitly incorporating kinematic topology to ensure compatibility with downstream applications.

Given an input depth image $\mathbf{I}$, an image encoder $\mathrm{Enc}(\cdot)$ first extracts dense visual features
\begin{equation}
  \bm{h}_\mathrm{img} = \mathrm{Enc}(\mathbf{I}),
\end{equation}
which encode both global context and local hand-part semantics.
In parallel, a 2D-to-3D projection module $\phi(\cdot)$ \cite{cheng2024handdagt,ren2023two} back-projects valid depth pixels into a 3D point cloud
\begin{equation}
  \bm{h}_{\mathrm{pcl}} = \phi(\mathbf{I}),
\end{equation}
providing explicit geometric cues that naturally accommodate occlusion, deformation, and structural absence.

To obtain robust and topology-aware representations, the encoder $\mathrm{Enc}(\cdot)$ is pre-trained using \textbf{Topology-Aware Distillation}, as shown in \cref{fig:framework-c}.
A teacher network trained on data with \emph{complete hand structures} provides reliable joint-level and relational priors, which are distilled into the student encoder.
The resulting encoder provides a topology-aware initialization and is subsequently fine-tuned together with the downstream pose modules while retaining the distillation objective.

To integrate appearance and geometry in a structure-aware manner, the image features are first aligned to the hand kinematic topology via a joint mapping function $\psi(\cdot)$,
\begin{equation}
  \psi(\bm{h}_\mathrm{img}) = \left\{ \bm{h}_{\mathrm{img}}^k \right\}_{k=1}^{K},
\end{equation}
while the point cloud features are processed by a multi-layer perceptron $\rho(\cdot)$ to obtain per-joint geometric descriptors
\begin{equation}
  \rho(\bm{h}_{\mathrm{pcl}}) = \left\{ \bm{h}_{\mathrm{pcl}}^k \right\}_{k=1}^{K}.
\end{equation}
Following \cite{ren2023two,cheng2024handdagt}, each point is augmented with the image feature sampled at its projected pixel, and $\rho(\cdot)$ obtains $\bm{h}_{\mathrm{pcl}}^k$ by encoding the $S=64$ points nearest to the current estimate of joint $k$, together with their relative offsets, with a shared MLP followed by max pooling; each stack re-centers these neighborhoods on the refined estimates.
We fuse these joint-aligned features into
\begin{equation}
  \bm{z} = \mathcal{F}\big(\{\bm{h}_{\mathrm{img}}^k\}_{k=1}^K, \{\bm{h}_{\mathrm{pcl}}^k\}_{k=1}^K\big),
\end{equation}
which jointly encodes appearance and kinematic structure.

The fused representation $\bm{z}$ is subsequently processed by \textbf{Progressive Kinematic Reasoning}, as illustrated in \cref{fig:framework-d}, to infer the full set of 3D hand joint coordinates:
\begin{equation}
  \hat{\mathcal{P}} = \mathcal{G}(\bm{z}),
\end{equation}
where $\mathcal{G}(\cdot)$ denotes the hierarchical root-to-tip inference process.
By explicitly modeling kinematic dependencies along the articulated hand structure, this module enables robust and topologically consistent pose estimation under incomplete configurations.
Details are presented in \cref{sec:r2t}.

Training follows the two-stage procedure in \cref{sec:training}. The student encoder is first pre-trained with $\mathcal{L}_\mathrm{distill}$ alone, after which downstream fine-tuning minimizes
\begin{equation} \label{eq:loss}
  \mathcal{L} = \mathcal{L}_\mathrm{pose}
  + \lambda_1\mathcal{L}_\mathrm{distill}
  + \lambda_2\mathcal{L}_\mathrm{root\text{-}tip}.
\end{equation}
Here, $\mathcal{L}_\mathrm{distill}$ preserves teacher guidance, $\mathcal{L}_\mathrm{pose}=K^{-1}\sum_{k=1}^{K}\|\hat{\bm p}_k-\bm p_k^*\|_2^2$ supervises joint regression, and $\mathcal{L}_\mathrm{root\text{-}tip}$ provides chain-balanced supervision.

\subsection{Topology-Aware Distillation}
\label{sec:h2i}

A key challenge in hand pose estimation under structural incompleteness is limited training data for incomplete kinematic structures.
Models trained mainly on intact hands lack structural priors for permanently absent joints and thus generalize poorly.
Topology-aware distillation (see \cref{fig:framework-c}) addresses this gap by transferring joint-level and relational priors from complete hands to structurally incomplete representations.

\myPara{Design Rationale.}
Although some kinematic elements may be absent, incomplete hands still follow the same underlying hand topology and functional constraints.
We leverage this observation by distilling topology-aware priors into the feature extractor, enabling anatomically plausible representations even when direct visual evidence is missing.
This is achieved through a teacher--student distillation framework that aligns both joint-level features and inter-joint relationships.

\myPara{Teacher--Student Feature Distillation.}
Distillation uses paired complete- and incomplete-hand samples, denoted by $\mathbf{I}_\mathrm{c}$ and $\mathbf{I}_\mathrm{i}$, respectively, which are processed by identical backbone architectures.
The complete-hand teacher $\mathrm{Tea}(\cdot)$ encodes anatomical priors, and the student encoder $\mathrm{Enc}(\cdot)$ transfers them to structurally incomplete inputs; the teacher remains fixed in both training stages.

For a paired sample $(\mathbf{I}_\mathrm{c}, \mathbf{I}_\mathrm{i})$, feature maps are extracted as
\begin{equation}
  \bm{h}_{\mathrm{c}}^{\mathrm{Tea}} = \mathrm{Tea}(\mathbf{I}_\mathrm{c}), \qquad
  \bm{h}_{\mathrm{i}}^{\mathrm{Enc}} = \mathrm{Enc}(\mathbf{I}_\mathrm{i}).
\end{equation}
Applying $\psi(\cdot)$ to the two feature maps yields the joint-level representations used for structural alignment:
\begin{equation}
  \psi(\bm{h}_{\mathrm{c}}^{\mathrm{Tea}}) = \{\bm{f}_{\mathrm{c}}^k\}_{k=1}^{K}, \qquad
  \psi(\bm{h}_{\mathrm{i}}^{\mathrm{Enc}}) = \{\bm{f}_{\mathrm{i}}^k\}_{k=1}^{K},
\end{equation}
where $\bm{f}_{\mathrm{c}}^k$ and $\bm{f}_{\mathrm{i}}^k$ are the teacher and student features for joint $k$, respectively; $K$ is the number of joints.

\myPara{Topological Relation Alignment.}
Relational knowledge distillation transfers structural information by preserving relationships among data examples~\cite{park2019relational}.
Following this relational transfer principle, we align inter-joint feature differences within paired complete- and incomplete-hand samples to preserve hand-specific structural relationships.
For any joint pair $(i,j)$, we define the relational differences as
$\Delta f_q^{ij} = f_q^i - f_q^j$, where $q \in \{c,i\}$.
We align the teacher and student relational representations by minimizing
\begin{equation}
  \mathcal{L}_{\mathrm{topo}} =
  \frac{1}{K^2} \sum_{i=1}^{K}\sum_{j=1}^{K}
  \left\|
  \Delta\bm{f}_{\mathrm{c}}^{ij}
  - \Delta\bm{f}_{\mathrm{i}}^{ij}
  \right\|_2^2 .
\end{equation}
With $\bm{r}_k=\bm{f}_{\mathrm{c}}^k-\bm{f}_{\mathrm{i}}^k$ and $\bar{\bm{r}}$ its mean over joints, this objective equals $\frac{2}{K}\sum_{k=1}^{K}\|\bm{r}_k-\bar{\bm{r}}\|_2^2=2\mathcal{L}_{\mathrm{joint}}-2\|\bar{\bm{r}}\|_2^2$ with $\mathcal{L}_{\mathrm{joint}}$ in \cref{eq:l_distill}, i.e., a centered feature-matching objective that discounts the offset shared by all joints and contains no explicit skeletal-edge or hierarchy weighting.
It therefore transfers the relative organization of joint features learned from intact hands, so the anatomical structure it conveys is implicit in the teacher features, whereas the explicit kinematic structure enters through progressive kinematic reasoning and $\mathcal{L}_{\mathrm{root\text{-}tip}}$.

\myPara{Distillation Objective.}
The overall distillation objective integrates global feature alignment, joint-level consistency, and topological relation preservation:
\begin{equation} \label{eq:l_distill}
  \mathcal{L}_{\mathrm{distill}} =
  \mathcal{L}_{\mathrm{global}} +
  \alpha \mathcal{L}_{\mathrm{joint}} +
  \beta \mathcal{L}_{\mathrm{topo}},
\end{equation}
where
$\mathcal{L}_{\mathrm{global}} = \|\bm{h}_{\mathrm{c}}^{\mathrm{Tea}} - \bm{h}_{\mathrm{i}}^{\mathrm{Enc}}\|_2^2$,
$\mathcal{L}_{\mathrm{joint}} = \frac{1}{K}\sum_{k=1}^{K} \|\bm{f}_{\mathrm{c}}^k - \bm{f}_{\mathrm{i}}^k\|_2^2$,
and $\mathcal{L}_{\mathrm{topo}}$ enforces relational consistency.
During pre-training, only the student encoder $\mathrm{Enc}(\cdot)$ is optimized.
The pre-trained parameters initialize the second stage, in which the student encoder continues to be optimized under $\mathcal{L}_\mathrm{distill}$ together with the pose-learning objectives.

\subsection{Progressive Kinematic Reasoning}
\label{sec:r2t}

Accurately estimating a complete hand pose under structural incompleteness requires explicit modeling of kinematic dependencies among joints.
Existing graph-based methods exploit hand topology mainly at the feature modeling stage, where joint features are aggregated and refined in a largely parallel manner.
Such feature-level reasoning is effective for capturing inter-joint correlations, but it does not explicitly constrain the order in which the final joint positions are inferred.
Under structural incompleteness, this distinction becomes important: distal joints often have weak or missing local observations, while their proximal ancestors are usually more observable and provide reliable anatomical anchors.
Therefore, predicting all joints in parallel may force the model to infer evidence-limited distal joints without sufficient kinematic guidance, as reflected in our ablation analysis (see \cref{tab:ablation}).
To address this limitation, we introduce a progressive kinematic reasoning module that organizes pose prediction along the anatomical root-to-tip chain, conditioning distal-joint estimation on parent predictions and joint-specific features.

\myPara{Design Rationale.}
The human hand forms a tree-structured kinematic graph, where the wrist serves as the root and each fingertip corresponds to a leaf node.
The spatial configuration of each joint is constrained by its ancestors along the kinematic chain.
Based on this structure, we formulate hand pose estimation as a progressive inference process that proceeds from the root toward distal joints.
This hierarchy provides an explicit kinematic prior for fixed-topology skeleton completion and complements feature-level graph reasoning.

\myPara{Hierarchical Pose Inference.}
Let $K$ denote the total number of hand joints and $\bm{z}$ denote the fused feature embedding described in the previous section.
The predicted joint coordinates are denoted as $\hat{\mathcal{P}} = \{\hat{\bm{p}}_1, \ldots, \hat{\bm{p}}_K\}$, where $\hat{\bm{p}}_k \in \mathbb{R}^3$.
The inference starts from the root joint (wrist, i.e., $k=1$), whose position is estimated directly from the global embedding:
\begin{equation}
  \hat{\bm{p}}_1 = \mathbf{W}_1 \bm{z} + \bm{b}_1,
\end{equation}
where $\mathbf{W}_1$ and $\bm{b}_1$ are learnable parameters.

For each subsequent joint $k > 1$, its position is inferred conditioned on its parent joint $\pi(k)$ and a joint-specific representation that integrates global and local information.
Specifically, we define a global--local interaction function
\begin{equation} \label{eq:gli}
  \bm{u}_k = \mathrm{GLI}_{\ell_k}(\bm{h}^k, \bm{z}),
\end{equation}
where $\bm{h}^k$ is the joint-aligned local feature, $\bm{z}$ provides global structural context, and $\ell_k$ denotes the kinematic level of joint $k$.
In our implementation, $\mathrm{GLI}_{\ell_k}(\cdot)$ is a lightweight concat--MLP fusion function. 
It concatenates the joint-aligned local feature $\bm{h}^k$ and the global structural feature $\bm{z}$, and maps the concatenated feature to $\bm{u}_k$ through a two-layer MLP.
The fused representation $\bm{u}_k$ is then combined with the parent prediction through a level-specific reasoning function:
\begin{equation} \label{eq:parent-reasoning}
  \hat{\bm{p}}_k = \mathcal{G}_{\ell_k}(\hat{\bm{p}}_{\pi(k)}, \bm{u}_k),
\end{equation}
where $\mathcal{G}_{\ell_k}(\cdot)$ is an MLP shared by all joints at the same kinematic level.
This level-wise parameter sharing reflects anatomical symmetry and promotes consistent inference across joints with similar functional roles in the hand.

\myPara{Training Objective.}
The progressive kinematic reasoning module is supervised using a root-to-tip pose loss organized according to the hierarchical structure of the hand skeleton.
Rather than aggregating all joints uniformly, the loss groups joint errors by finger kinematic chain and provides balanced supervision from the wrist to distal joints.
Let $\mathcal{C}$ denote the set of root-to-tip kinematic chains corresponding to individual fingers,
where each chain $c \in \mathcal{C}$ is an ordered sequence of joints from the root to a fingertip.
Averaging the joint errors over all root-to-tip chains gives
\begin{equation}
  \mathcal{L}_{\mathrm{root\text{-}tip}} =
  \frac{1}{|\mathcal{C}|}
  \sum_{c \in \mathcal{C}}
  \frac{1}{|c|}
  \sum_{k \in c}
  \left\| \hat{\bm{p}}_k - \bm{p}_k^{*} \right\|_2^2,
\end{equation}
where $\bm{p}_k^{*}$ denotes the ground-truth position of the $k$-th joint.

By organizing supervision along root-to-tip kinematic chains, this auxiliary loss aligns the grouping of joint errors with the progressive inference order.
The parent-conditioned update in \cref{eq:parent-reasoning} provides the explicit kinematic dependency, while the loss ensures that proximal and distal predictions along every finger chain receive supervision.

\myPara{Discussion.}
By explicitly encoding kinematic hierarchy and conditioning each joint prediction on its parent and global context, the proposed reasoning module avoids flat joint regression and enforces anatomically meaningful constraints.
This progressive formulation enables complete and anatomically consistent pose estimation under severe structural variations and complements the topology-aware distillation module by performing structure-consistent inference at the pose level.

\subsection{Training Procedure}
\label{sec:training}

InsHaPE follows the two-stage strategy in \cref{algo:training}: Stage 1 learns structural features, and Stage 2 jointly adapts them with the pose modules to incomplete inputs.

\myPara{Stage 1: Topology-Aware Distillation Pre-training.}
In the first stage, a teacher network is trained on the complete-hand NYU training set to capture comprehensive anatomical and kinematic representations.
A student encoder is then optimized using paired complete and incomplete hand images, supervised to align with the teacher's joint-level features and topological relations by minimizing the distillation loss.
During this stage, only the student encoder is updated.
The resulting parameters initialize the downstream fine-tuning stage.

\myPara{Stage 2: Downstream Fine-tuning.}
In the second stage, the student encoder, feature-fusion module, and progressive reasoning module are jointly fine-tuned.
Paired complete/incomplete inputs retain topology-aware distillation during fine-tuning, while the incomplete input is also used for pose regression and root-to-tip supervision.
This retains the teacher's structural guidance while adapting the downstream model to structurally incomplete poses.

\begin{algorithm}[t]
  \SetAlgoLined
  \KwIn{Paired complete/incomplete-hand data $\mathcal{D}_\mathrm{c},\mathcal{D}_\mathrm{i}$}
  \KwOut{Trained InsHaPE model}

  \tcp{Stage 1: Distillation Pre-training}
  Train $\mathrm{Tea}(\cdot)$ on $\mathcal{D}_\mathrm{c}$ for complete-hand pose regression\;
  \For{each paired mini-batch $(\mathbf{I}_\mathrm{c}, \mathbf{I}_\mathrm{i})$}{
  $\bm{h}_\mathrm{c} = \mathrm{Tea}(\mathbf{I}_\mathrm{c})$, $\bm{h}_\mathrm{i} = \mathrm{Enc}(\mathbf{I}_\mathrm{i})$\;
  Map features to joints: $\{\bm{f}_{\mathrm{c}}^k\}$, $\{\bm{f}_{\mathrm{i}}^k\}$\;
  Compute $\mathcal{L}_\mathrm{distill}$ and update $\mathrm{Enc}(\cdot)$\;
  }
  \tcp{Stage 2: Downstream Fine-tuning}
  \For{each paired mini-batch $(\mathbf{I}_\mathrm{c},\mathbf{I}_\mathrm{i})$}{
  $\bm{h}_\mathrm{c} = \mathrm{Tea}(\mathbf{I}_\mathrm{c})$, $\bm{h}_\mathrm{i} = \mathrm{Enc}(\mathbf{I}_\mathrm{i})$\;
  Map features to joints: $\{\bm{f}_{\mathrm{c}}^k\}$, $\{\bm{f}_{\mathrm{i}}^k\}$\;
  Extract joint-level point-cloud features $\{\bm{h}_{\mathrm{pcl}}^k\}$\;
  Fuse features: $\bm{z} = \mathcal{F}(\{\bm{f}_\mathrm{i}^k\}, \{\bm{h}_{\mathrm{pcl}}^k\})$\;
  Predict joint coordinates: $\{\hat{\bm{p}}_k\} = \mathcal{G}(\bm{z})$\;
  Compute the fine-tuning objective $\mathcal{L}$ in \cref{eq:loss}\;
  Update $\mathrm{Enc}(\cdot)$, fusion, and reasoning modules\;
  }
  \caption{Training procedure of InsHaPE}
  \label{algo:training}
\end{algorithm}

\section{Theoretical Analysis}
\label{sec:theory}

This section provides analytical intuition rather than an unconditional performance guarantee.
It characterizes why \emph{structural incompleteness} can amplify relational ambiguity and error propagation and identifies sufficient conditions under which the proposed structural priors can mitigate these effects.
We therefore provide a conditional characterization of the structural inductive biases introduced by InsHaPE.
Recall that topology-aware distillation first initializes the student representation and then remains active as a structural regularizer in the fine-tuning objective in \cref{eq:loss}.
Let
$\widehat{\mathcal{R}}(f)=\frac{1}{n}\sum_{i=1}^n \ell\big(f(\mathbf{I}_i),\mathbf{y}_i\big)$
denote the empirical risk, where $\ell$ is $L$-Lipschitz and uniformly bounded by $B$.
Let $f_\lambda$ denote a model learned with the two structural objectives, with $\lambda$ collectively denoting their weights, and let $f_0$ denote the corresponding model learned without these constraints.

\begin{theorem}[Inductive Bias under Structural Incompleteness]
  \label{thm:inductive_bias_full}
  For any $\delta\in(0,1)$, with probability at least $1-\delta$,
  \begin{align}
    \mathcal{R}(f_\lambda) & \le \widehat{\mathcal{R}}(f_\lambda)+\Gamma+\Delta+\kappa+C_0, \\
    \mathcal{R}(f_0)       & \le \widehat{\mathcal{R}}(f_0)+\Gamma_0+D_0+\kappa_0+C_0',
  \end{align}
  where $\mathcal{R}(\cdot)$ denotes the expected risk and $C_0,C_0'\ge0$ collect the remaining approximation and noise terms. Under the stated regularity and local contraction conditions, the following hold:
  \begin{itemize}
    \item $\Gamma = 2L\,\mathfrak{R}_n(\mathcal{H}(\rho_1,\rho_2))
            + 3B\sqrt{\log(2/\delta)/(2n)}$ and $\Gamma_0=2L\,\mathfrak{R}_n(\mathcal{H}_0)+3B\sqrt{\log(2/\delta)/(2n)}$,
          where $\mathcal{H}(\rho_1,\rho_2)=\{f\in\mathcal{H}_0:\mathcal{L}_{\mathrm{distill}}(f)\le\rho_1,\ \mathcal{L}_{\mathrm{root\text{-}tip}}(f)\le\rho_2\}$; hence, $\Gamma\le \Gamma_0$.
    \item $\Delta = C_1\sqrt{\rho_1}$ upper-bounds the relational discrepancy induced by structural incompleteness under the assumption that the relational projector is Lipschitz and its output is bounded, while $D_0$ denotes the corresponding unconstrained discrepancy term.
    \item $\kappa = C_2\gamma^{d_{\min}}$ characterizes propagated perturbations along kinematic chains when the learned local updates satisfy the sufficient contraction condition $\gamma\in(0,1)$, whereas $\kappa_0$ is its unconstrained counterpart.
  \end{itemize}
  Accordingly, the theorem identifies a sufficient regime in which the proposed constraints reduce hypothesis complexity, relational discrepancy, and kinematic error propagation.
\end{theorem}

\begin{proof}
  Since the constrained hypothesis class $\mathcal{H}(\rho_1,\rho_2)$ satisfies $\mathcal{H}(\rho_1,\rho_2)\subseteq\mathcal{H}_0$, its empirical Rademacher complexity obeys $\mathfrak{R}_n(\mathcal{H}(\rho_1,\rho_2))\le \mathfrak{R}_n(\mathcal{H}_0)$. By the standard generalization bound for $L$-Lipschitz and $B$-bounded losses, with probability at least $1-\delta$, any $f$ in a hypothesis class $\mathcal{F}$ satisfies $\mathcal{R}(f)\le \widehat{\mathcal{R}}(f)+2L\,\mathfrak{R}_n(\mathcal{F})+3B\sqrt{\log(2/\delta)/(2n)}$. Applying this bound to $\mathcal{F}=\mathcal{H}(\rho_1,\rho_2)$ yields the term $\Gamma$, while applying it to $\mathcal{F}=\mathcal{H}_0$ yields $\Gamma_0$, hence $\Gamma\le\Gamma_0$. We note that this complexity relation characterizes the structural inductive bias introduced by the constrained hypothesis class rather than, by itself, guaranteeing a lower expected risk.

  Under structural incompleteness, multiple complete poses may correspond to the same partial observation, leading to ambiguity in the latent representations of absent joints. Using the joint features defined in \cref{sec:h2i}, let $\Delta\bm f_{\mathrm c}^{ij}=\bm f_{\mathrm c}^{i}-\bm f_{\mathrm c}^{j}$ and $\Delta\bm f_{\mathrm i}^{ij}=\bm f_{\mathrm i}^{i}-\bm f_{\mathrm i}^{j}$. Assuming that the joint-feature projector $\psi$ is Lipschitz and its outputs are bounded, the discrepancy between the complete- and incomplete-input relational distributions can be upper-bounded by a constant multiple of the average relational alignment error, for example through a linear-kernel MMD argument. Since $\beta\mathcal{L}_{\mathrm{topo}}\le\mathcal{L}_{\mathrm{distill}}$ for $\beta>0$, this gives $\mathrm{Disc}_{\mathrm{rel}}\le \widetilde C_1\sqrt{\mathcal{L}_{\mathrm{distill}}(f)}$, where $\widetilde C_1$ absorbs the factor $1/\sqrt{\beta}$. For notational simplicity, the theorem denotes this constant by $C_1$. Hence, for a constrained solution satisfying $\mathcal{L}_{\mathrm{distill}}(f_\lambda)\le\rho_1$, we obtain $\mathrm{Disc}_{\mathrm{rel}}(f_\lambda)\le C_1\sqrt{\rho_1}=\Delta$.

  Consider the progressive root-to-tip inference along the kinematic tree. For each edge $e=(u\rightarrow v)$, let the local update map be $g_e$ so that $\hat{\bm{p}}_v=g_e(\hat{\bm{p}}_u,\bm{u}_v)$.
If the learned local update is contractive with respect to the parent state, namely $\|\partial g_e/\partial \hat{\bm{p}}_u\|_2\le\gamma<1$, unrolling this recursion along any root-to-joint path of length $d$ yields geometric decay of propagated perturbations, so the contribution of upstream error to a distal joint is bounded by $C_2\gamma^{d}$. Taking the minimum depth $d_{\min}$ over structurally absent distal joints gives $\kappa=C_2\gamma^{d_{\min}}$. Importantly, $\gamma<1$ is introduced here as a sufficient local contraction condition rather than as a guaranteed property of every model or input. Under this condition, progressive parent-conditioned reasoning provides a mechanism for attenuating upstream perturbations along the kinematic hierarchy.

  Combining these results yields the stated conditional characterization with terms $(\Gamma,\Delta,\kappa)$ for $f_\lambda$ and $(\Gamma_0,D_0,\kappa_0)$ for the corresponding unconstrained formulation. The result therefore describes how the proposed structural priors can reduce effective hypothesis complexity, constrain relational ambiguity, and attenuate error propagation when the stated sufficient conditions hold. This concludes the proof.
\end{proof}

\myPara{Discussion.}
The analysis separates three effects tied to structural incompleteness.
Topology-aware distillation introduces a structural inductive bias and constrains relational ambiguity by aligning joint relations learned from complete hand structures, while progressive kinematic reasoning introduces explicit parent-conditioned dependencies along the anatomical hierarchy.
Under the sufficient local contraction condition, this hierarchical formulation attenuates the propagation of proximal perturbations toward evidence-limited distal joints.
We emphasize that these results provide a conditional analytical explanation of the robustness of InsHaPE rather than an unconditional guarantee of performance improvement.

\section{Experiments} \label{sec:exp}
We begin by introducing the experimental setup.
We then compare InsHaPE with state-of-the-art methods on both standard and structurally incomplete benchmarks, followed by cross-setting generalization studies.
Finally, we present detailed ablation studies and supplementary analyses to examine the contribution of each component and to assess the computational efficiency and robustness of InsHaPE.

\subsection{Experimental Setting}
\label{sec:exp_setting}

\myPara{External Datasets.}
In addition to NYU and IncNYU described in \cref{sec:problem}, we use the MSRA dataset~\cite{sun2015cascaded} and construct its structurally incomplete counterpart, IncMSRA. Following the IncNYU construction pipeline, we select anatomically valid regions, remove the corresponding hand segments from distal to proximal, and render the modified hand under the original acquisition configuration. Each incomplete sample retains its complete counterpart and joint annotations. Following the standard MSRA protocol, we use leave-one-subject-out cross-validation over its nine subjects, average over all nine folds, and follow its output and evaluation joint protocol. 

\myPara{Evaluation Metrics.}
We evaluate hand pose estimation performance using the mean per-joint position error (MPJPE), measured in millimeters, under three complementary metrics.
For a test set of $N$ samples, let $\mathcal{K}_{\mathrm{eval}}$ denote the benchmark-specific evaluation joint set, with $K_{\mathrm{eval}}=|\mathcal{K}_{\mathrm{eval}}|$. For each sample $n$, let $\mathcal{S}_{\text{com}}^{(n)}$ and $\mathcal{S}_{\text{inc}}^{(n)}$ denote the joints associated with complete and structurally incomplete regions, respectively. We define $\mathcal{N}_{q}=\{n:|\mathcal{S}_{q}^{(n)}|>0\}$ for $q\in\{\text{com},\text{inc}\}$.
We compute the three evaluation metrics over these sets as follows:
\begin{align}
  E_{\text{com}} & =
  \frac{1}{|\mathcal{N}_{\text{com}}|} \sum_{n\in\mathcal{N}_{\text{com}}}
  \frac{1}{|\mathcal{S}_{\text{com}}^{(n)}|}
  \sum_{k \in \mathcal{S}_{\text{com}}^{(n)}}
  \left\|
  \hat{\bm{p}}_{k}^{(n)} - \bm{p}_{k}^{*(n)}
  \right\|_2,        \\
  E_{\text{inc}} & =
  \frac{1}{|\mathcal{N}_{\text{inc}}|} \sum_{n\in\mathcal{N}_{\text{inc}}}
  \frac{1}{|\mathcal{S}_{\text{inc}}^{(n)}|}
  \sum_{k \in \mathcal{S}_{\text{inc}}^{(n)}}
  \left\|
  \hat{\bm{p}}_{k}^{(n)} - \bm{p}_{k}^{*(n)}
  \right\|_2,        \\
  E_{\text{avg}} & =
  \frac{1}{N} \sum_{n=1}^{N}
  \frac{1}{K_{\mathrm{eval}}}
  \sum_{k\in\mathcal{K}_{\mathrm{eval}}}
  \left\|
  \hat{\bm{p}}_{k}^{(n)} - \bm{p}_{k}^{*(n)}
  \right\|_2,
\end{align}
where $\hat{\bm{p}}_{k}^{(n)}$ and $\bm{p}_{k}^{*(n)}$ denote the predicted and reference positions of the $k$-th joint in the $n$-th sample. A region-specific metric is reported only when its corresponding sample set $\mathcal{N}_{q}$ is nonempty.
Specifically, $E_{\text{com}}$ measures MPJPE over anatomically intact regions, reflecting the model's accuracy on normally observed joints.
$E_{\text{inc}}$ measures MPJPE over joints associated with structurally incomplete regions.
For synthetic incomplete-hand benchmarks constructed from complete-hand annotations, the hidden complete-hand joints provide a controlled reference for evaluating whether a method can infer anatomically plausible and fixed-topology-compatible placeholders for missing regions. 
Thus, $E_{\text{inc}}$ should be interpreted as a proxy for structured skeleton completion under persistent absence, rather than as exact recovery of non-existent anatomy in real incomplete hands.
$E_{\text{avg}}$ complements $E_{\text{com}}$ and $E_{\text{inc}}$ by reporting MPJPE over all joints in $\mathcal{K}_{\mathrm{eval}}$, providing an overall assessment of fixed-topology hand pose estimation.
For real incomplete-hand data, hidden complete-hand references are unavailable; we therefore evaluate downstream interaction rather than phantom-joint errors (see \cref{sec:system}).

\myPara{Implementation Details.}
For all experiments, we use ResNet-50 as the backbone for both the teacher and student encoders, following common practice in hand pose estimation. The encoder output is mapped to joint-wise features via a lightweight MLP with two hidden layers of 256 and 128 units, respectively. The point cloud branch adopts a three-layer MLP with hidden dimensions 128, 128, and 64. The feature fusion function $\mathcal{F}$ concatenates image and point cloud features for each joint, followed by a linear projection. The progressive kinematic reasoning module is implemented as a set of MLPs, with one MLP shared across all joints at each kinematic level. Each MLP consists of two hidden layers with 128 units and ReLU activations.
For the NYU-based experiments, the model predicts a fixed $K=23$ joint representation, while the reported NYU and IncNYU metrics use the standard 14-joint evaluation subset. For MSRA, MPJPE is computed over all 21 annotated joints following the standard evaluation protocol~\cite{sun2015cascaded}. We set the stage-2 distillation weight $\lambda_1=0.01$, the stage-2 chain-loss weight $\lambda_2=0.05$, and the internal distillation weights $\alpha=0.01$ and $\beta=0.01$, following the sensitivity analysis in \cref{tab:loss}. The models are trained with the Adam optimizer with a learning rate of $0.01$, a batch size of 64, and a weight decay of $0.01$. The student encoder is pre-trained for 50 epochs using paired complete and incomplete hand images. It is then fine-tuned together with the fusion and reasoning modules for 100 epochs, with topology-aware distillation retained during this stage. All experiments are conducted on a single NVIDIA RTX 4090 GPU using PyTorch.
We augment the data with random rotation ($\pm 30^\circ$), scaling ($\pm 10\%$), and occlusion. Depth images are resized to $128 \times 128$ pixels, with 1024 sampled points per hand.

\begin{table}[t]
  \centering
  \caption{
    Performance comparison with state-of-the-art methods on the \textbf{NYU} dataset. The \colorbox{cyan!10}{cyan cell} marks the best result. Input modalities include 2D depth images (D), 3D voxels (V), and 3D point clouds (P).
  }
  \label{tab:comparison-nyu}
  \begin{tabular}{rrcr}
    \toprule
    \textbf{Method}                             & \textbf{Venue} & \textbf{Input} & \textbf{MPJPE}           \\
    \midrule
    Ren-9x6x6~\cite{guo2017towards}             & JVCI'18        & D              & 12.69                    \\
    DeepPrior++~\cite{oberweger2017deepprior++} & ICCV'17        & D              & 12.24                    \\
    Pose-Ren~\cite{chen2020pose}                & NEUCOM'20      & D              & 11.81                    \\
    DenseReg~\cite{wan2018dense}                & CVPR'18        & D              & 10.20                    \\
    CrossInfoNet~\cite{du2019crossinfonet}      & CVPR'19        & D              & 10.08                    \\
    A2J \cite{xiong2019a2j}                     & ICCV'19        & D              & 8.61                     \\
    AWR \cite{huang2020awr}                     & AAAI'20        & D              & 7.48                     \\
    TriHorn-Net \cite{rezaei2023trihorn}        & ESWA'23        & D              & 7.68                     \\
    JGR-P2O~\cite{fang2020jgr}                  & ECCV'20        & D              & 8.29                     \\
    SSRN~\cite{ren2021spatial}                  & NEUCOM'21      & D              & 7.37                     \\
    PHG~\cite{ren2021pose}                      & TOC'21         & D              & 7.39                     \\
    HandPointNet~\cite{ge2018hand}              & CVPR'18        & P              & 10.54                    \\
    Hand-Transformer~\cite{huang2020hand}       & ECCV'20        & P              & 9.80                     \\
    Point-to-Point~\cite{ge2018point}           & ECCV'18        & P              & 9.10                     \\
    V2V~\cite{moon2018v2v}                      & CVPR'18        & V              & 8.42                     \\
    HandFolding~\cite{cheng2021handfoldingnet}  & ICCV'21        & P              & 8.58                     \\
    HandR2N2~\cite{cheng2023handr2n2}           & ICCV'23        & P              & 7.27                     \\
    IPNet~\cite{ren2023two}                     & AAAI'23        & D+P            & 7.17                     \\
    HandDiff \cite{cheng2024handdiff}           & CVPR'24        & D+P            & 7.38                     \\
    HandDAGT \cite{cheng2024handdagt}           & ECCV'24        & D+P            & 7.12                     \\
    \hdashline
    InsHaPE (Ours)                              & --             & D+P            & \cellcolor{cyan!10} 7.07 \\
    \bottomrule
  \end{tabular}
\end{table}

\subsection{Comparison with State-of-the-Art Methods}

The results in \cref{tab:comparison-nyu,tab:comparison-incnyu} show that InsHaPE achieves the lowest overall error on NYU and IncNYU, demonstrating its effectiveness for complete and structurally incomplete hands.

\myPara{Performance on NYU.}
We first evaluate InsHaPE on the standard NYU dataset to verify that introducing inclusivity-oriented design does not compromise performance under conventional complete-hand settings. As shown in \cref{tab:comparison-nyu}, InsHaPE achieves an MPJPE of 7.07\,mm, comparable to the strongest recent baseline HandDAGT (7.12\,mm).
This result indicates that the proposed design preserves competitive accuracy on fully articulated hands while primarily improving robustness under structural incompleteness.
The small performance difference on NYU is expected because the dataset contains only complete hand structures, where strong geometric and appearance cues are available for all joints. In this regime, most recent methods, especially those leveraging multimodal inputs or diffusion-based modeling, already approach the performance ceiling. Importantly, InsHaPE does not sacrifice accuracy on intact hands despite being explicitly designed to handle structural incompleteness, confirming that its inductive biases are compatible with standard pose estimation.

\myPara{Performance on IncNYU.}
The advantage of InsHaPE becomes significantly more pronounced on the IncNYU benchmark, which explicitly evaluates robustness under structural incompleteness. When all methods are retrained on IncNYU (bottom part of \cref{tab:comparison-incnyu}), InsHaPE establishes a clear new state of the art, achieving an average error $E_{\text{avg}}$ of 8.00. Compared to the strongest baseline HandDAGT, InsHaPE reduces $E_{\text{com}}$, $E_{\text{inc}}$, and $E_{\text{avg}}$ by 32.1\%, 29.9\%, and 30.4\%, respectively.
The baselines in this comparison are trained without paired complete-hand observations; under the same paired supervision, HandDAGT reaches an $E_{\text{avg}}$ of 8.18 (ID~3 in \cref{tab:ablation}), compared with 8.00 for InsHaPE.
Notably, while most baselines exhibit a substantial performance gap between intact joints ($E_{\text{com}}$) and incomplete joints ($E_{\text{inc}}$), InsHaPE significantly narrows this gap. This indicates that the model improves not only overall accuracy but also fixed-topology skeleton completion for structurally absent regions, the central challenge considered throughout this work.

\begin{table}[t]
  \centering
  \setlength{\tabcolsep}{0.5em}
  \caption{
    Performance comparison with state-of-the-art methods on the {\textbf{IncNYU}} dataset. A \colorbox{cyan!10}{cyan cell} marks the best result in each column. Input modalities include 2D depth images (D), 3D voxels (V), and 3D point clouds (P).
  }
  \resizebox{\linewidth}{!}{
    \begin{tabular}{rrcrrr}
      \toprule
      \multirow{2.5}{*}{\bf Method}        & \multirow{2.5}{*}{\bf Venue} & \multirow{2.5}{*}{\bf Input} & \multicolumn{3}{c}{\bf w/o IncNYU retraining}                                                         \\
      \cmidrule(lr){4-6}
                                           &                              &                              & $E_{\text{com}}$                              & $E_{\text{inc}}$          & $E_{\text{avg}}$          \\
      \midrule
      A2J \cite{xiong2019a2j}              & ICCV'19                      & D                            & 82.58                                         & 91.85                     & 87.28                     \\
      AWR \cite{huang2020awr}              & AAAI'20                      & D                            & 67.44                                         & 94.16                     & 73.72                     \\
      TriHorn-Net \cite{rezaei2023trihorn} & ESWA'23                      & D                            & 15.59                                         & 46.79                     & 22.07                     \\
      HandR2N2 \cite{cheng2023handr2n2}    & ICCV'23                      & P                            & 16.25                                         & \cellcolor{cyan!10}26.36  & 19.23                     \\
      IPNet \cite{ren2023two}              & AAAI'23                      & D+P                          & 20.49                                         & 35.36                     & 23.65                     \\
      HandDiff \cite{cheng2024handdiff}    & CVPR'24                      & D+P                          & 15.17                                         & 38.22                     & 20.20                     \\
      HandDAGT \cite{cheng2024handdagt}    & ECCV'24                      & D+P                          & \cellcolor{cyan!10} 13.28                     & 34.53                     & 18.47                     \\
      \hdashline
      InsHaPE (Ours)                       & --                           & D+P                          & \cellcolor{cyan!10} 13.28                     & 33.40                     & \cellcolor{cyan!10} 17.44 \\
      \midrule
                                           &                              &                              & \multicolumn{3}{c}{\bf w/ IncNYU retraining}                                                          \\
      \cmidrule(lr){4-6}
      A2J \cite{xiong2019a2j}              & ICCV'19                      & D                            & 15.54                                         & 31.83                     & 20.16                     \\
      AWR \cite{huang2020awr}              & AAAI'20                      & D                            & 15.51                                         & 33.37                     & 19.83                     \\
      TriHorn-Net \cite{rezaei2023trihorn} & ESWA'23                      & D                            & 11.22                                         & 33.12                     & 15.67                     \\
      HandR2N2 \cite{cheng2023handr2n2}    & ICCV'23                      & P                            & 10.76                                         & 22.56                     & 14.08                     \\
      IPNet \cite{ren2023two}              & AAAI'23                      & D+P                          & 11.42                                         & 15.87                     & 12.36                     \\
      HandDiff \cite{cheng2024handdiff}    & CVPR'24                      & D+P                          & 13.76                                         & 25.57                     & 15.99                     \\
      HandDAGT \cite{cheng2024handdagt}    & ECCV'24                      & D+P                          & 8.54                                          & 19.61                     & 11.49                     \\ \hdashline
      InsHaPE (Ours)                       & --                           & D+P                          & \cellcolor{cyan!10} 5.80                      & \cellcolor{cyan!10} 13.75 & \cellcolor{cyan!10} 8.00  \\
      \bottomrule
    \end{tabular}
  }
  \label{tab:comparison-incnyu}
\end{table}

\myPara{Generalization from NYU to IncNYU.}
We further evaluate direct generalization by applying models trained on NYU to IncNYU without retraining (top part of \cref{tab:comparison-incnyu}). In this setting, all models, including InsHaPE, are trained on complete NYU data only and see no synthetic incomplete observations in any training stage; hence TAD is not applied.
This setting is particularly challenging, as models are forced to extrapolate from complete-hand data to structurally incomplete configurations.
Under this protocol, the baselines suffer substantial performance degradation, especially on incomplete joints, with $E_{\text{inc}}$ often exceeding 30\,mm. InsHaPE achieves the lowest overall error ($E_{\text{avg}}=17.44$), improving on HandDAGT's 18.47 by 5.58\%, while its incomplete-region error ($E_{\text{inc}}=33.40$) remains higher than the best incomplete-region result of 26.36 obtained by HandR2N2. These results support improved overall robustness to the NYU-to-IncNYU shift without implying superiority on every joint subset.

\myPara{Method-Level Analysis.}
The performance gains are consistent with the complementary roles of InsHaPE's two components. First, the topology-aware distillation mechanism transfers relational priors from complete hand structures, reducing ambiguity in the latent representation of missing joints and mitigating the domain shift between intact and incomplete hands. This is particularly critical in IncNYU, where multiple anatomically plausible completions may correspond to the same partial observation.
Second, the progressive kinematic reasoning module explicitly encodes hierarchical kinematic dependencies and constrains error propagation along the hand skeleton. By conditioning distal joint predictions on proximal estimates, this module mainly reduces errors on incomplete joints in the controlled ablations (\cref{tab:ablation,tab:parent_control}); since the compared methods differ from InsHaPE in several components, the gains over graph-based methods such as HandDAGT reflect the full framework rather than this module alone.
Together, these designs enable InsHaPE to achieve superior accuracy on standard benchmarks while substantially improving robustness and generalization under structural incompleteness, validating its suitability for inclusive MR applications.

\myPara{Training Stability.}
To assess the robustness of InsHaPE against random initialization, we further conduct experiments with three different random seeds on IncNYU.
The model achieves consistent performance across runs, with mean errors of 5.83\,$\pm$\,0.04, 13.55\,$\pm$\,0.35, and 8.01\,$\pm$\,0.10 for $E_{\mathrm{com}}$, $E_{\mathrm{inc}}$, and $E_{\mathrm{avg}}$, respectively.
The averaged results are consistent with the main comparison in \cref{tab:comparison-incnyu}, indicating that the reported performance is stable across different training runs.

\begin{table}[t]
    \centering
    \MStabcaption{tab:msra_inc}{Performance comparison on MSRA and IncMSRA. We report MPJPE (mm) on standard MSRA, direct generalization from MSRA to IncMSRA without re-training, and performance after re-training on IncMSRA.}
    \setlength{\tabcolsep}{3.5pt}
    \resizebox{\linewidth}{!}{
    \begin{tabular}{lccccc}
    \toprule
    \bf Method & \bf Venue & \bf Input &\bf  MSRA & \bf  \makecell{MSRA$\rightarrow$\\IncMSRA} & \bf  IncMSRA \\
    \midrule
    IPNet~\cite{ren2023two}
    & AAAI'23 & D+P & 6.64 & 7.18 & 7.36 \\
    
    HandDiff~\cite{cheng2024handdiff}
    & CVPR'24 & D+P & 6.53 & 6.90 & 5.89 \\
    
    HandDAGT~\cite{cheng2024handdagt}
    & ECCV'24 & D+P & 6.38 & 6.79 & 6.74 \\
    InsHaPE (Ours)
    & -- & D+P & \cellcolor{cyan!10}\textbf{5.52} & \cellcolor{cyan!10}\textbf{6.23} & \cellcolor{cyan!10}\textbf{5.77} \\
    \bottomrule
    \end{tabular}
    }
\end{table}

\myPara{External Validation on MSRA and IncMSRA.}
We further evaluate InsHaPE on the MSRA benchmark and its structurally incomplete variant, IncMSRA. As shown in \cref{tab:msra_inc}, InsHaPE achieves the lowest MPJPE across all three evaluation settings. On the standard MSRA dataset, InsHaPE obtains an error of 5.52\,mm, outperforming HandDAGT, HandDiff, and IPNet by 13.5\%, 15.5\%, and 16.9\%, respectively. Under the more challenging direct generalization setting from MSRA to IncMSRA without re-training, InsHaPE maintains a clear advantage with an MPJPE of 6.23\,mm, compared with 6.79\,mm for HandDAGT, 6.90\,mm for HandDiff, and 7.18\,mm for IPNet, corresponding to relative reductions of 8.2\%, 9.7\%, and 13.2\%, respectively. After re-training on IncMSRA, InsHaPE achieves 5.77\,mm, improving upon the next-best result of 5.89\,mm from HandDiff by 2.0\%. These results further demonstrate that InsHaPE generalizes beyond NYU/IncNYU while retaining robustness to structural incompleteness.

\subsection{Computational Comparison with State-of-the-Art Methods}

\myPara{Full Model Comparison.}
We evaluate InsHaPE (w/ 3 stacks) against recent state-of-the-art methods that use both depth images and point clouds (D+P) in terms of computational efficiency.
The results are summarized in \cref{tab:computation}.
Despite achieving the lowest average error ($E_{\text{avg}}=8.00$), InsHaPE also has the lowest measured model-inference latency among the compared D+P methods, with an average runtime of 29.43\,ms per frame on an NVIDIA RTX 4090.
This is substantially faster than HandDAGT (45.47\,ms) and IPNet (45.65\,ms). These measurements concern model inference only and do not include sensor acquisition, data transfer, downstream recognition, or rendering latency.
In terms of model complexity, InsHaPE contains more parameters (108.01M) than the compared baselines.
However, its computational cost in FLOPs (23.47G) is on par with HandDAGT (23.74G) and significantly lower than HandDiff (46.68G).
This suggests that the additional parameters do not translate proportionally into FLOPs or measured model-inference latency.
Importantly, the larger model yields a substantial performance gain: InsHaPE reduces the average error by 30.4\% compared to HandDAGT on IncNYU while achieving faster model inference on the tested hardware.
This offers a favorable trade-off, but end-to-end wearable MR latency remains untested.

\begin{table}[!t]
  \centering
  \setlength{\tabcolsep}{5pt}
  \caption{Computational comparison on the IncNYU dataset.}
  \label{tab:computation}
  \begin{tabular}{l S[table-format=3.2] S[table-format=2.2] S[table-format=2.2] S[table-format=2.2]}
    \toprule
    \textbf{Method}                      & {\makecell{\textbf{Params} \\ \textbf{(M)}}} & {\makecell{\textbf{FLOPs} \\ \textbf{(G)}}} & {\makecell{\textbf{Inference} \\ \textbf{(ms)}}} & {$E_{\text{avg}}$} \\
    \midrule
    IPNet \cite{ren2023two}              & 39.51                     & 2.97  & 45.65 & 12.36 \\
    HandDiff \cite{cheng2024handdiff}    & 67.55                     & 46.68 & 59.75 & 15.99 \\
    HandDAGT \cite{cheng2024handdagt}    & 69.48                     & 23.74 & 45.47 & 11.49 \\ \hdashline
    InsHaPE w/ 1 stack\phantom{s} (Ours) & 74.20                     & 17.09 & 18.12 & 10.67 \\
    InsHaPE w/ 2 stacks (Ours)           & 91.10                     & 20.42 & 23.44 & 8.10  \\
    InsHaPE w/ 3 stacks (Ours)           & 108.01                    & 23.47 & 29.43 & 8.00  \\
    \bottomrule
  \end{tabular}
\end{table}

\myPara{Stack-Depth and Efficiency Analysis.}
Similar to HandDAGT~\cite{cheng2024handdagt}, InsHaPE employs stacked progressive kinematic reasoning blocks to iteratively refine hand pose predictions and enforce structural consistency.
Each stack takes the joint coordinates and joint features predicted by the preceding stack as input, and the first stack starts from the initial estimates regressed from the global image and point-cloud codes.
While deeper stacking can improve representational capacity, it also increases computational cost, which may limit deployment on latency-sensitive MR systems.
To analyze this trade-off, we evaluate InsHaPE with one, two, and three reasoning stacks.
As shown in \cref{tab:computation}, increasing the number of stacks consistently improves accuracy, reducing $E_{\text{avg}}$ from 10.67 (1 stack) to 8.10 (2 stacks), and finally to 8.00 (3 stacks).
This improvement comes at the cost of higher computation, with inference time increasing from 18.12\,ms to 29.43\,ms and FLOPs from 17.09G to 23.47G, indicating diminishing accuracy gains beyond two stacks.
Importantly, even with a single reasoning stack, InsHaPE already outperforms all compared state-of-the-art methods.
Specifically, InsHaPE with one stack achieves an average error of 10.67, which is substantially lower than HandDAGT (11.49), IPNet (12.36), and HandDiff (15.99), while operating at less than half of HandDAGT's inference latency.
Thus, both proposed modules remain effective with a single stack.

\subsection{Ablation Study}

\myPara{Impact of Core Modules.}
We first analyze the contribution of each core component in InsHaPE, as summarized in \cref{tab:ablation}.
Removing the global--local interaction (GLI, ID~2) increases $E_{\text{avg}}$ from 8.00 to 8.15 in a single run, suggesting that explicitly fusing joint-level local cues with global structural context is beneficial.
This is consistent with GLI balancing fine-grained appearance cues and holistic hand structure.
For ID~3, we replace the progressive root-to-tip reasoning module with a parallel prediction head following the common flat joint prediction paradigm \cite{cheng2024handdagt}.
This head takes the same fused features as input and predicts all joint offsets in parallel, without kinematic-level decomposition or parent-conditioned inference.
Since all configurations share the same three-stack design and ID~3 also disables GLI, the gap between IDs~1 and~3 reflects the joint effect of the two components.
Between IDs~2 and~3, where GLI is absent in both, PKR* lowers $E_{\text{inc}}$ from 14.15 to 13.88 at a small cost on complete joints; as this single-run difference lies within the seed variation of the full model, the seed-matched control in \cref{tab:parent_control} provides the main evidence for parent conditioning.
Without TAD, the effect of PKR* is much larger: IDs~5 and~6 differ only in PKR*, which lowers $E_{\text{inc}}$ from 19.61 to 16.31 and $E_{\text{avg}}$ from 11.49 to 10.95.
The smaller gain once TAD is applied is consistent with both components reducing the ambiguity of evidence-limited distal joints.
This pattern is consistent with the motivation of progressive kinematic reasoning, as distal joints under structural incompleteness have limited local observations and can be constrained by more reliable proximal cues.
Excluding topology-aware distillation (IDs~4 and~5) leads to a clear degradation of 20--37\% in $E_{\text{avg}}$, demonstrating its role in narrowing the domain gap between complete and incomplete hand structures.
Notably, removing distillation while retaining progressive reasoning (ID~4) already causes a substantial increase in $E_{\text{com}}$, indicating that the learned representation itself becomes less reliable without transferred structural priors.
Finally, when all three components are removed (ID~6), the average error increases dramatically by 44\%.
This severe performance drop highlights that InsHaPE does not rely on any single module; rather, robust performance emerges from the synergy between topology-aware distillation, GLI, and progressive kinematic reasoning.

\begin{table}[t]
  \centering
  \setlength{\tabcolsep}{0.5em}
  \caption{Core module ablation results on IncNYU.
    Reported percentages indicate relative error changes compared to the full model (ID~1).
    \textbf{TAD}: Topology-Aware Distillation (\S\ref{sec:h2i});
    \textbf{PKR*}: Progressive Kinematic Reasoning (\S\ref{sec:r2t}) without Global--Local Interaction (GLI, \cref{eq:gli}).}
  \label{tab:ablation}
  \resizebox{\linewidth}{!}{
    \begin{tabular}{cccclll}
      \toprule
      \multirow{2.5}{*}{\textbf{ID}} & \multicolumn{3}{c}{\textbf{Configuration}} & \multirow{2.5}{*}{$E_{\text{com}}$} & \multirow{2.5}{*}{$E_{\text{inc}}$} & \multirow{2.5}{*}{$E_{\text{avg}}$}                                                                 \\
      \cmidrule(lr){2-4}
                                     & TAD                                        & PKR*                                & GLI                                                                                                                                       \\
      \midrule
      1                              & \ding{52}                                  & \ding{52}                           & \ding{52}                           & 5.80                                & 13.75                    & \phantom{0}8.00                    \\ \hdashline
      2                              & \ding{52}                                  & \ding{52}                           & \ding{55}                           & 5.93 $^{\uparrow 2\%}$              & 13.88 $^{\uparrow 1\%}$  & \phantom{0}8.15 $^{\uparrow 2\%}$  \\
      3                              & \ding{52}                                  & \ding{55}                           & \ding{55}                           & 5.77$^{\approx 0\%}$                & 14.15 $^{\uparrow 3\%}$  & \phantom{0}8.18 $^{\uparrow 2\%}$  \\
      4                              & \ding{55}                                  & \ding{52}                           & \ding{52}                           & 6.94 $^{\uparrow 20\%}$             & 13.84 $^{\uparrow 1\%}$  & \phantom{0}9.56 $^{\uparrow 20\%}$ \\
      5                              & \ding{55}                                  & \ding{52}                           & \ding{55}                           & 8.16 $^{\uparrow 41\%}$             & 16.31 $^{\uparrow 19\%}$ & 10.95 $^{\uparrow 37\%}$           \\
      6                              & \ding{55}                                  & \ding{55}                           & \ding{55}                           & 8.54 $^{\uparrow 47\%}$             & 19.61 $^{\uparrow 43\%}$ & 11.49 $^{\uparrow 44\%}$           \\
      \bottomrule
    \end{tabular}
  }
\end{table}

\begin{table}[!t]
  \centering
  \setlength{\tabcolsep}{0.5em}
  \caption{Distillation loss ablation results on IncNYU.
    Reported percentages are error changes compared to ID~1.
  }
  \label{tab:feature_ablation}
  \resizebox{\linewidth}{!}{
    \begin{tabular}{cccclll}
      \toprule
      \multirow{2.5}{*}{\textbf{ID}} & \multicolumn{3}{c}{\textbf{Configuration}} & \multirow{2.5}{*}{$E_{\text{com}}$} & \multirow{2.5}{*}{$E_{\text{inc}}$} & \multirow{2.5}{*}{$E_{\text{avg}}$}                                                                 \\
      \cmidrule(lr){2-4}
                                     & $\mathcal{L}_{\mathrm{global}}$            & $\mathcal{L}_{\mathrm{joint}}$      & $\mathcal{L}_{\mathrm{topo}}$                                                                                                             \\
      \midrule
      1                              & \ding{52}                                  & \ding{52}                           & \ding{52}                           & 5.80                                & 13.75                    & \phantom{0}8.00                    \\ \hdashline
      2                              & \ding{52}                                  & \ding{52}                           & \ding{55}                           & 6.72 $^{\uparrow 16\%}$             & 14.20 $^{\uparrow 3\%}$  & \phantom{0}8.81 $^{\uparrow 10\%}$ \\
      3                              & \ding{52}                                  & \ding{55}                           & \ding{55}                           & 7.66 $^{\uparrow 32\%}$             & 16.69 $^{\uparrow 21\%}$ & 10.27 $^{\uparrow 28\%}$           \\
      4                              & \ding{55}                                  & \ding{52}                           & \ding{52}                           & 5.97 $^{\uparrow 3\%}$              & 14.00 $^{\uparrow 2\%}$  & \phantom{0}8.27 $^{\uparrow 3\%}$  \\
      5                              & \ding{55}                                  & \ding{52}                           & \ding{55}                           & 6.19 $^{\uparrow 7\%}$              & 14.10 $^{\uparrow 3\%}$  & \phantom{0}8.41 $^{\uparrow 5\%}$  \\
      6                              & \ding{55}                                  & \ding{55}                           & \ding{52}                           & 6.37 $^{\uparrow 10\%}$             & 14.23 $^{\uparrow 3\%}$  & \phantom{0}8.62 $^{\uparrow 8\%}$  \\
      7                              & \ding{55}                                  & \ding{55}                           & \ding{55}                           & 6.94 $^{\uparrow 20\%}$             & 13.84 $^{\uparrow 1\%}$  & \phantom{0}9.56 $^{\uparrow 20\%}$ \\
      \bottomrule
    \end{tabular}
  }
\end{table}

\myPara{Impact of Distillation Losses.}
We further investigate the contribution of each distillation loss, with results reported in \cref{tab:feature_ablation}.
The full combination $\mathcal{L}_{\mathrm{global}} + \mathcal{L}_{\mathrm{joint}} + \mathcal{L}_{\mathrm{topo}}$ (ID~1, see \cref{eq:l_distill}) achieves the lowest errors across all metrics.
Removing topological alignment while retaining global and joint-level alignment (ID~2) increases $E_{\mathrm{avg}}$ from 8.00 to 8.81.
Removing global alignment while retaining joint-level and topological alignment (ID~4) increases it to 8.27.
The benefit of joint-level alignment is supported by two controlled comparisons: adding it to global-only distillation reduces $E_{\mathrm{avg}}$ from 10.27 to 8.81 (IDs~3 and~2), while adding it to topology-only distillation reduces the error from 8.62 to 8.27 (IDs~6 and~4).
The single-loss configurations (IDs~3, 5, and~6) all yield higher overall errors than the full combination, while removing all distillation losses (ID~7) increases $E_{\mathrm{avg}}$ to 9.56.
Together, these results support the complementary benefits of global, joint-level, and topological alignment.

\myPara{Impact of Kinematic Modeling Paradigms.}
To analyze the contribution of progressive kinematic reasoning, we compare three paradigms (see \cref{fig:method-cmp}): flat joint-wise modeling, cascaded finger-wise modeling, and hierarchical root-to-tip reasoning.
All three variants share the same three-stack design, pose regression head, objective (see \cref{eq:loss}), and training schedule, and contain 107.95M, 114.57M, and 108.01M parameters, respectively; they differ in how joint relations are modeled within each stack.
All variants retain the feature-level graph reasoning of HandDAGT~\cite{cheng2024handdagt}: the flat variant corresponds to ID~3 in \cref{tab:ablation}, i.e., HandDAGT with TAD, the cascaded variant models each finger with a separate module, and only the hierarchical variant, i.e., the full model, uses GLI and parent conditioning.
As shown in \cref{fig:kinematic-ablation}, the hierarchical formulation achieves the best overall performance, reducing $E_{\mathrm{avg}}$ to 8.00, compared with 8.18 for flat modeling and 8.50 for cascaded modeling.
Compared with cascaded modeling, hierarchical root-to-tip reasoning reduces $E_{\mathrm{com}}$, $E_{\mathrm{inc}}$, and $E_{\mathrm{avg}}$ by 5.23\%, 3.85\%, and 5.88\%, respectively; as shown below, this gap is not explained by GLI alone.
Although flat modeling achieves a slightly lower $E_{\mathrm{com}}$ (5.77 vs. 5.80), it does not provide explicit structural reasoning and results in higher overall error.
These results suggest that the gain of PKR cannot be explained solely by introducing additional prediction stages.
Because GLI is used only in the hierarchical variant, \cref{fig:kinematic-ablation} compares the prediction architectures as a whole.
When GLI is also removed from the hierarchical model (ID~2 in \cref{tab:ablation}), it still yields the lowest $E_{\mathrm{inc}}$ of the three architectures (13.88 vs.\ 14.15 and 14.30) and a lower $E_{\mathrm{avg}}$ than the cascaded variant (8.15 vs.\ 8.50), with $E_{\mathrm{avg}}$ close to the flat variant (8.18) in this single run.

\begin{figure}
    \centering
    \includegraphics[width=\linewidth]{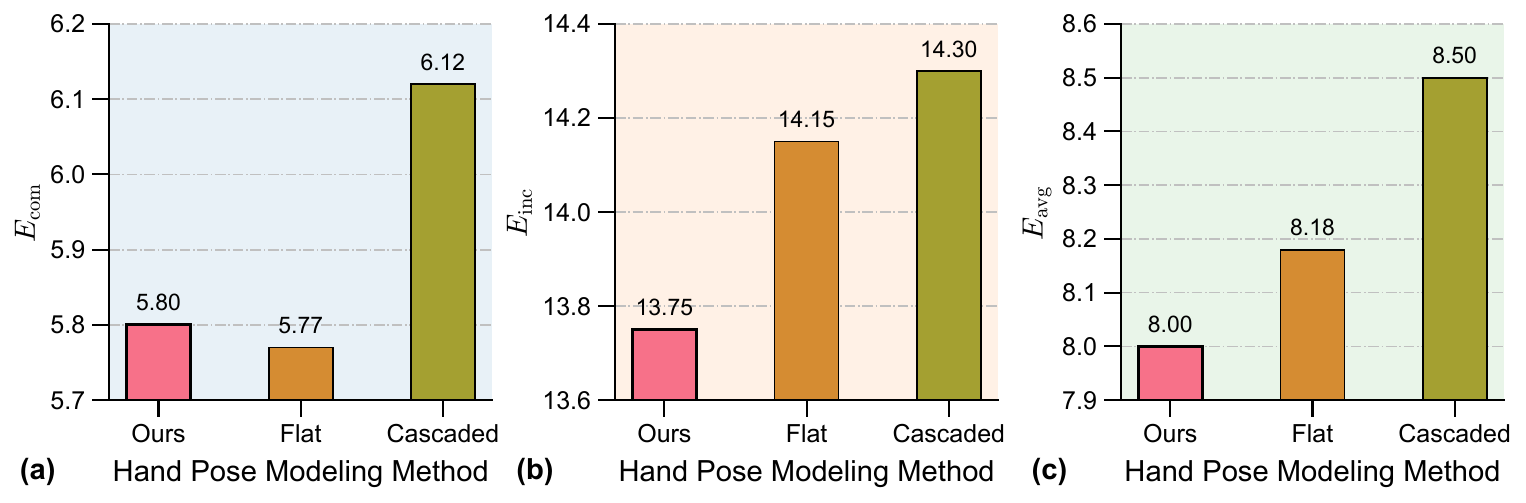}
    \MSfigcaption{fig:kinematic-ablation}{
    Ablation results of different hand pose modeling paradigms (see \cref{fig:method-cmp}).
    We compare flat joint-wise modeling, cascaded finger-wise modeling, and the proposed hierarchical root-to-tip reasoning in terms of (a) $E_{\mathrm{com}}$, (b) $E_{\mathrm{inc}}$, and (c) $E_{\mathrm{avg}}$.
    }{}
\end{figure}

\myPara{Parent-Conditioning Control.}
To isolate the role of parent conditioning, we retrain the full model with the parent prediction $\hat{\bm{p}}_{\pi(k)}$ in \cref{eq:parent-reasoning} set to zero, keeping all other components and training settings unchanged, and use the same three random seeds for both models.
Because the model is retrained with this constant input, the input carries no information about the parent prediction, so the control removes parent conditioning while keeping the architecture unchanged.
As shown in \cref{tab:parent_control}, removing the parent input mainly affects incomplete joints, raising $E_{\mathrm{inc}}$ in each of the three seeds and from 13.55 to 14.08 on average, a difference larger than the seed-to-seed standard deviation of either model, while $E_{\mathrm{com}}$ is almost unchanged; the full-model result in \cref{tab:ablation} is one of these three runs.
This is consistent with the intended role of parent conditioning: proximal estimates matter most when distal joints lack direct visual observations.

\begin{table}[t]
\centering
\MStabcaption{tab:parent_control}{
Parent-conditioning control on IncNYU (mean $\pm$ standard deviation over three matched random seeds).
}
\resizebox{\linewidth}{!}{%
\begin{tabular}{lccc}
\toprule
\bf Configuration & $E_{\mathrm{com}}$ & $E_{\mathrm{inc}}$ & $E_{\mathrm{avg}}$ \\
\midrule
InsHaPE (Full model) & 5.83\,$\pm$\,0.04 & 13.55\,$\pm$\,0.35 & 8.01\,$\pm$\,0.10 \\
w/o Parent Conditioning & 5.86\,$\pm$\,0.05 & 14.08\,$\pm$\,0.28 & 8.13\,$\pm$\,0.08 \\
\bottomrule
\end{tabular}%
}
\end{table}

\begin{table}[t]
\centering
\MStabcaption{tab:pointcloud_ablation}{
Geometric point-cloud ablation on IncNYU.
}{}
\begin{tabular}{llll}
\toprule
\bf Configuration
& $E_{\mathrm{com}}$
& $E_{\mathrm{inc}}$
& $E_{\mathrm{avg}}$ \\
\midrule
InsHaPE (Full model)
& 5.80
& 13.75
& \phantom{0}8.00 \\
w/o Point Cloud
& 9.34 $^{\uparrow61\%}$
& 16.71 $^{\uparrow22\%}$
& 11.35 $^{\uparrow42\%}$ \\
\bottomrule
\end{tabular}
\end{table}

\myPara{Impact of Geometric Point-Cloud Features.}
To evaluate the contribution of geometric point-cloud features, we remove the point-cloud branch while keeping other components unchanged.
Here, the point cloud is derived from the input depth image through depth-to-3D projection.
As shown in \cref{tab:pointcloud_ablation}, removing the geometric representation increases the overall error from 8.00 to 11.35, with consistent degradation on both complete and incomplete joints.
These results demonstrate that the 3D geometric representation derived from depth provides complementary structural information beyond depth appearance features, which is important for robust pose estimation under structural incompleteness.
Despite the performance drop, the point-cloud-free variant remains competitive with existing depth-only methods in \cref{tab:comparison-incnyu}, indicating that the proposed framework effectively exploits complementary appearance and geometric cues from depth inputs.

\begin{figure}[!t]
  \centering
  \includegraphics[width=\linewidth]{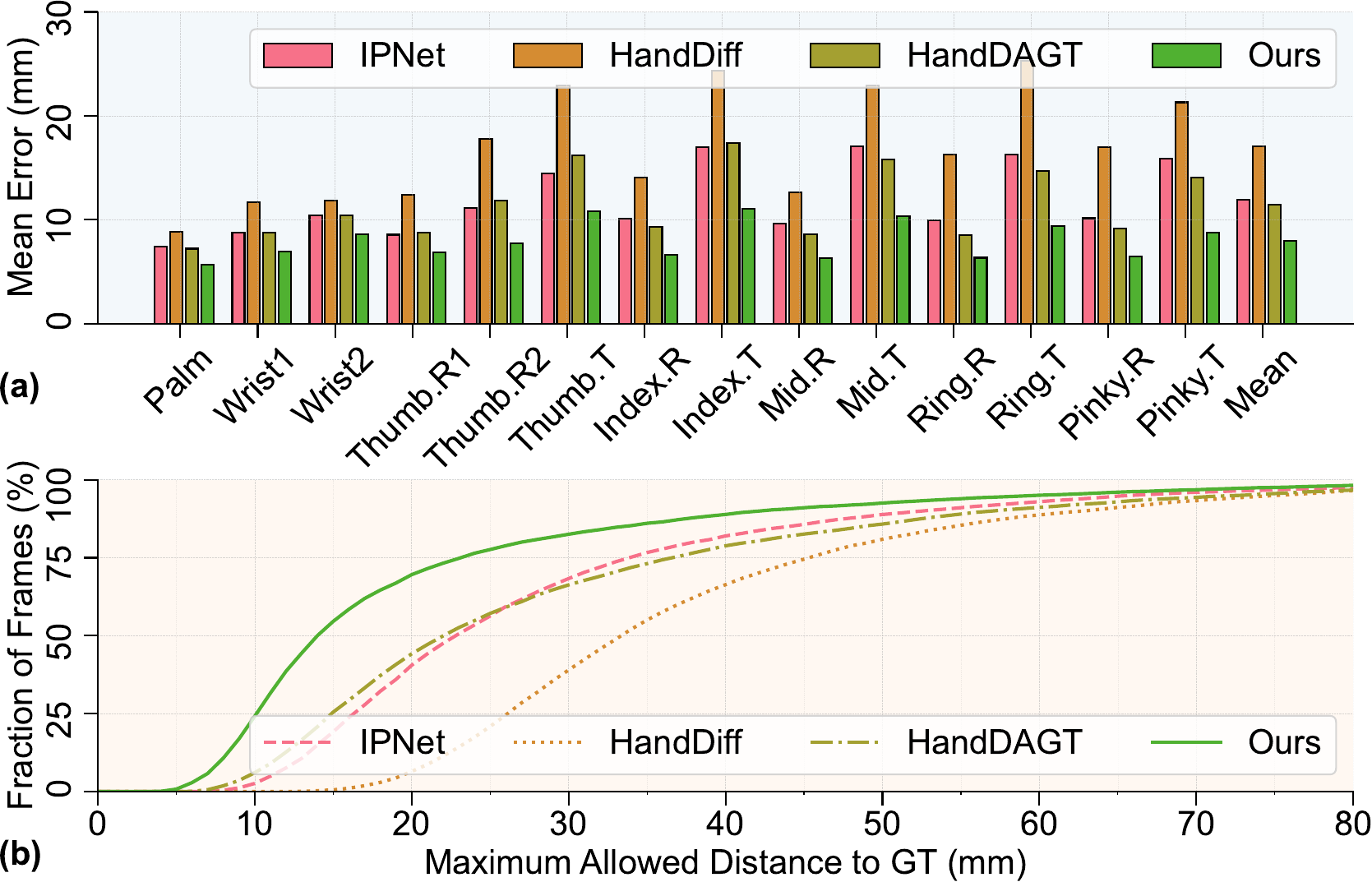}
  \caption{Joint-wise error analysis on the IncNYU dataset.
    \subref{fig:nyu_error-a}: Mean per-joint position error (MPJPE), showing increasing difficulty from palm to fingertips.
    \subref{fig:nyu_error-b}: PCK-style curves illustrating the fraction of frames within a given distance threshold to the ground truth (GT).}
  \label{fig:nyu_error}{
    \MSphantomsubcaptions{%
      \phantomsubcaption\label{fig:nyu_error-a}%
      \phantomsubcaption\label{fig:nyu_error-b}%
    }%
  }
\end{figure}

\begin{table}[t]
\centering
\MStabcaption{tab:pairing_ablation}{
Comparison under paired supervision on IncNYU.
}{}
\resizebox{\linewidth}{!}{%
\begin{tabular}{l r@{\,}l r@{\,}l r@{\,}l}
\toprule
\bf Configuration
& \multicolumn{1}{c@{\,}}{$E_{\mathrm{com}}$} &
& \multicolumn{1}{c@{\,}}{$E_{\mathrm{inc}}$} &
& \multicolumn{1}{c@{\,}}{$E_{\mathrm{avg}}$} & \\
\midrule
HandDiff
& 13.76 &
& 25.57 &
& 15.99 & \\
HandDiff + Paired Supervision
& 9.87 & $^{\downarrow28\%}$
& 16.93 & $^{\downarrow34\%}$
& 12.29 & $^{\downarrow23\%}$ \\
InsHaPE (Ours)
& \textbf{5.80} &
& \textbf{13.75} &
& \textbf{8.00} & \\
\bottomrule
\end{tabular}%
}
\end{table}

\myPara{Impact of Paired Complete-Hand Supervision.}
To examine the benefit of access to paired complete-hand information, we additionally adapt HandDiff~\cite{cheng2024handdiff} using the same paired complete/incomplete training data available to InsHaPE.
Specifically, we apply paired feature distillation to its feature extraction module using the corresponding complete-hand observations, and then initialize HandDiff with the distilled feature extractor for subsequent training.
This provides HandDiff with the same access to paired complete-hand information while preserving its original pose estimation framework.
As shown in \cref{tab:pairing_ablation}, this configuration reduces HandDiff's $E_{\mathrm{avg}}$ from 15.99 to 12.29, indicating that access to paired complete-hand information is beneficial.
InsHaPE achieves a lower overall error of 8.00 under the evaluated settings.
This comparison controls access to paired training data, while the methods differ in how they exploit that information.
Unlike InsHaPE, this HandDiff variant uses the distilled feature extractor only for initialization and does not retain distillation during fine-tuning.
A HandDAGT control that matches InsHaPE in this respect is given by IDs~6 and~3 in \cref{tab:ablation}: ID~6 corresponds to HandDAGT, and ID~3 adds TAD retained during fine-tuning, reducing $E_{\mathrm{avg}}$ from 11.49 to 8.18 on IncNYU.

\begin{table}[t]
  \centering
  \caption{Sensitivity analysis of loss-related hyperparameters on IncNYU.
    We report $E_{\mathrm{avg}}$ under different values of each hyperparameter,
    while keeping the other hyperparameters fixed at their default settings.
    The best result in each row is highlighted in bold.}
  \label{tab:loss}
  \setlength{\tabcolsep}{5.5pt}
  \renewcommand{\arraystretch}{1.08}
  \begin{tabular}{lcccc}
    \toprule
    \textbf{Hyperparameter} & \textbf{0.005} & \textbf{0.01} & \textbf{0.05} & \textbf{0.1} \\
    \midrule
    $\lambda_1$             & 8.14           & \textbf{8.00} & 8.29          & 8.47         \\
    $\lambda_2$             & 8.18           & 8.12          & \textbf{8.00} & 8.32         \\
    $\alpha$                & 8.07           & \textbf{8.00} & 8.14          & 8.18         \\
    $\beta$                 & 8.11           & \textbf{8.00} & 8.15          & 8.23         \\
    \bottomrule
  \end{tabular}
\end{table}

\myPara{Impact of Loss Weights.}
We analyze the sensitivity of InsHaPE to loss-related hyperparameters, including the stage-2 distillation weight ($\lambda_1$), the stage-2 chain-loss weight ($\lambda_2$), and the internal balancing factors $\alpha$ and $\beta$ within the distillation objective.
As shown in \cref{tab:loss}, InsHaPE remains robust across the tested hyperparameter values.
The best performance occurs at $\lambda_1=0.01$, with only marginal degradation at smaller or larger values, indicating that topology-aware distillation does not require precise tuning.
Performance is similarly robust to $\lambda_2$, with an optimum around $\lambda_2=0.05$, suggesting that progressive kinematic reasoning acts as a structural regularizer rather than a dominant loss term.
The internal weights $\alpha$ and $\beta$, which balance joint-level and topological alignment, also have little effect: performance remains near the optimum even when they vary by an order of magnitude.
Overall, these results indicate InsHaPE requires little loss-weight tuning.

\subsection{Quantitative and Qualitative Results}

\myPara{Joint-Wise Error Statistics.}
\cref{fig:nyu_error} provides the results of the joint-wise error distribution on IncNYU.
As shown in \cref{fig:nyu_error-a}, errors are lowest at the palm center and gradually increase toward distal joints, especially the fingertips.
This trend is expected, as distal joints exhibit higher degrees of freedom and larger variability, making them more challenging to estimate accurately.
Notably, InsHaPE consistently maintains lower errors across all joints, with a more controlled error growth toward the fingertips, indicating its robustness under structural incompleteness.
In \cref{fig:nyu_error-b}, we report the Percentage of Correct Keypoints (PCK)-style curves, showing the fraction of frames whose joint errors fall within varying distance thresholds.
The larger area under the curve achieved by InsHaPE demonstrates consistently higher accuracy across a wide range of tolerances.
This result indicates that InsHaPE not only improves mean accuracy but also yields fewer large errors, particularly for challenging distal joints.

\begin{figure}[t]
    \centering
    \sf
    \footnotesize
    \setlength{\tabcolsep}{0pt}
    \resizebox{\linewidth}{!}{
    \begin{tabular}{:>{\centering\arraybackslash}m{1.5em}:c:c:c:c:c:}
    \hdashline
    \cellcolor{gray!15} \# & \cellcolor{green!15} GT & 
         \cellcolor{yellow!15} Ours & 
         \cellcolor{orange!15} HandDAGT & 
         \cellcolor{violet!15} HandDiff & 
         \cellcolor{teal!15} IPNet \\ \hdashline
    \cellcolor{gray!15} \rotatebox{90}{0479} &
    \imgplain{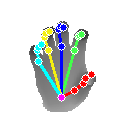}  & 
    \imgcirc{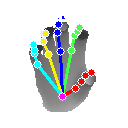}{55}{25}{6}  & 
    \imgcirc{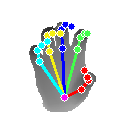}{55}{25}{6}  & 
    \imgcirc{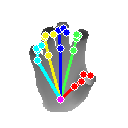}{55}{25}{6} &
    \imgcirc{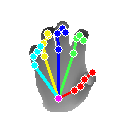}{55}{25}{6} \\ \hdashline
    
    \cellcolor{gray!15} \rotatebox[origin=c]{90}{2099} &
    \imgplain{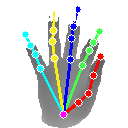}  & 
    \imgcirc{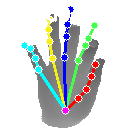}{50}{49}{8}  & 
    \imgcirc{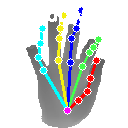}{50}{49}{8}   & 
    \imgcirc{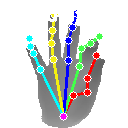}{50}{49}{8}  &
    \imgcirc{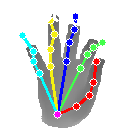}{50}{49}{8}    \\  \hdashline
    \cellcolor{gray!15} \rotatebox[origin=c]{90}{5423} &
    \imgplain{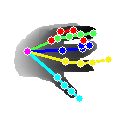}  & 
    \imgcirc{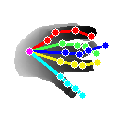}{33}{10}{8}  & 
    \imgcirc{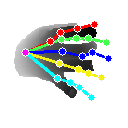}{33}{10}{8}  & 
    \imgcirc{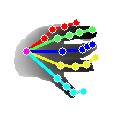}{33}{10}{8} &
    \imgcirc{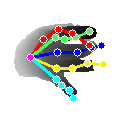}{33}{10}{8}   \\ 
    \hdashline
    \end{tabular}
    }
    \MSfigcaption{fig:qualitative_results}{Qualitative comparisons of InsHaPE with state-of-the-art baselines on the IncNYU dataset. Circles mark highlighted parts.}{}
\end{figure}

\begin{figure}[!t]
    \centering
    \includegraphics[width=\linewidth]{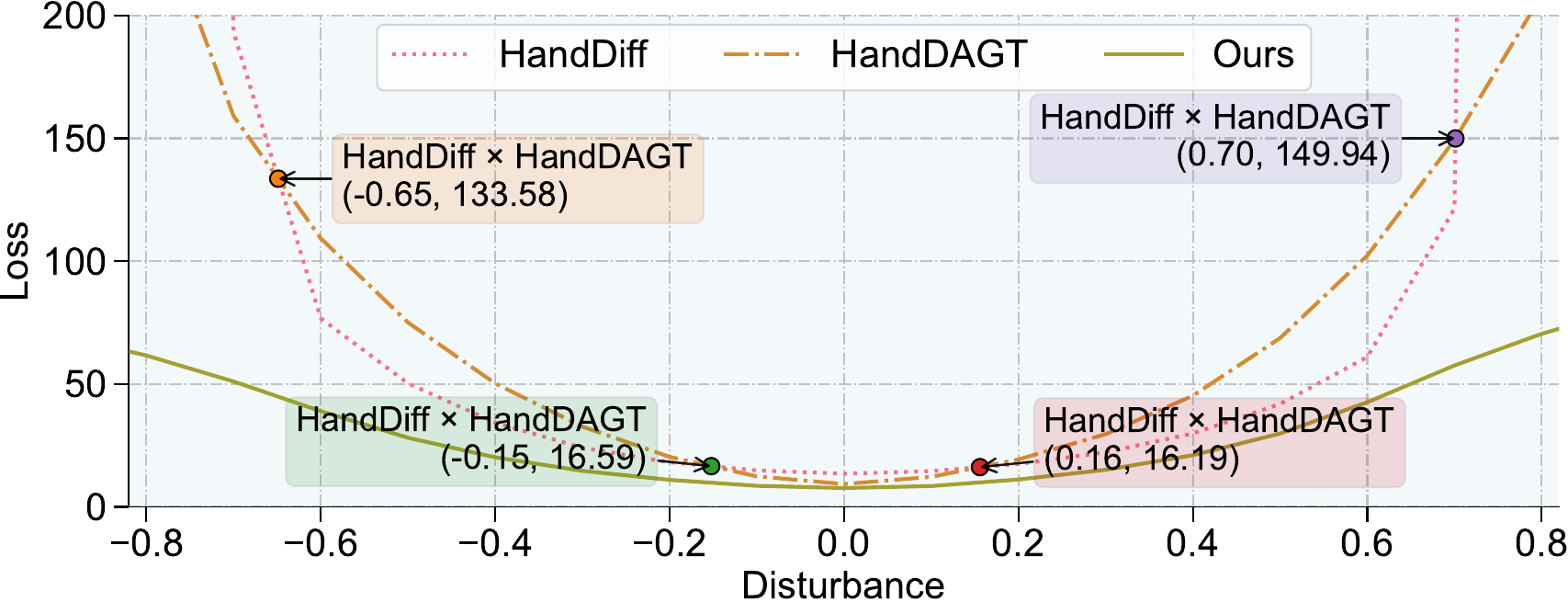}
    \MSfigcaption{fig:loss_landscape}{One-dimensional loss-landscape slices on IncNYU. InsHaPE exhibits a flatter loss profile than the compared methods along the sampled, filter-normalized perturbation directions.}{}
\end{figure}

\myPara{Qualitative Comparisons.}
\cref{fig:qualitative_results} presents qualitative comparisons between InsHaPE and representative state-of-the-art methods, including HandDAGT, HandDiff, and IPNet, on the IncNYU dataset.
Baseline methods frequently fail in regions affected by amputations or severe structural incompleteness, leading to missing joints, distorted finger configurations, or kinematically implausible poses.
In contrast, InsHaPE produces anatomically coherent, fixed-topology pose representations more consistently.
In particular, InsHaPE better localizes distal joints such as fingertips and preserves global hand structure, even when intermediate joints are absent.
These visual results align with the quantitative improvements and highlight the effectiveness of topology-aware distillation and progressive kinematic reasoning in handling structurally incomplete hands.

\myPara{Loss Landscape Analysis.}
To examine the local sensitivity of the trained models to parameter perturbations, we visualize averaged one-dimensional loss profiles on IncNYU using the filter normalization technique in \cite{li2018visualizing}.
For each trained model, we independently sample 10 random perturbation directions and apply filter normalization.
We evaluate the loss along each direction as the perturbation coefficient varies over $[-0.8,0.8]$ and average the losses across the 10 directions at each coefficient.
As shown in \cref{fig:loss_landscape}, InsHaPE exhibits a flatter average loss profile than HandDAGT and HandDiff, suggesting lower sensitivity to the sampled parameter perturbations.
Although averaging over multiple directions reduces dependence on any single direction, these profiles remain a qualitative assessment of local parameter sensitivity.
They do not guarantee generalization or verify the contraction condition in \cref{sec:theory}.

\begin{figure}
    \centering
    \includegraphics[width=\linewidth]{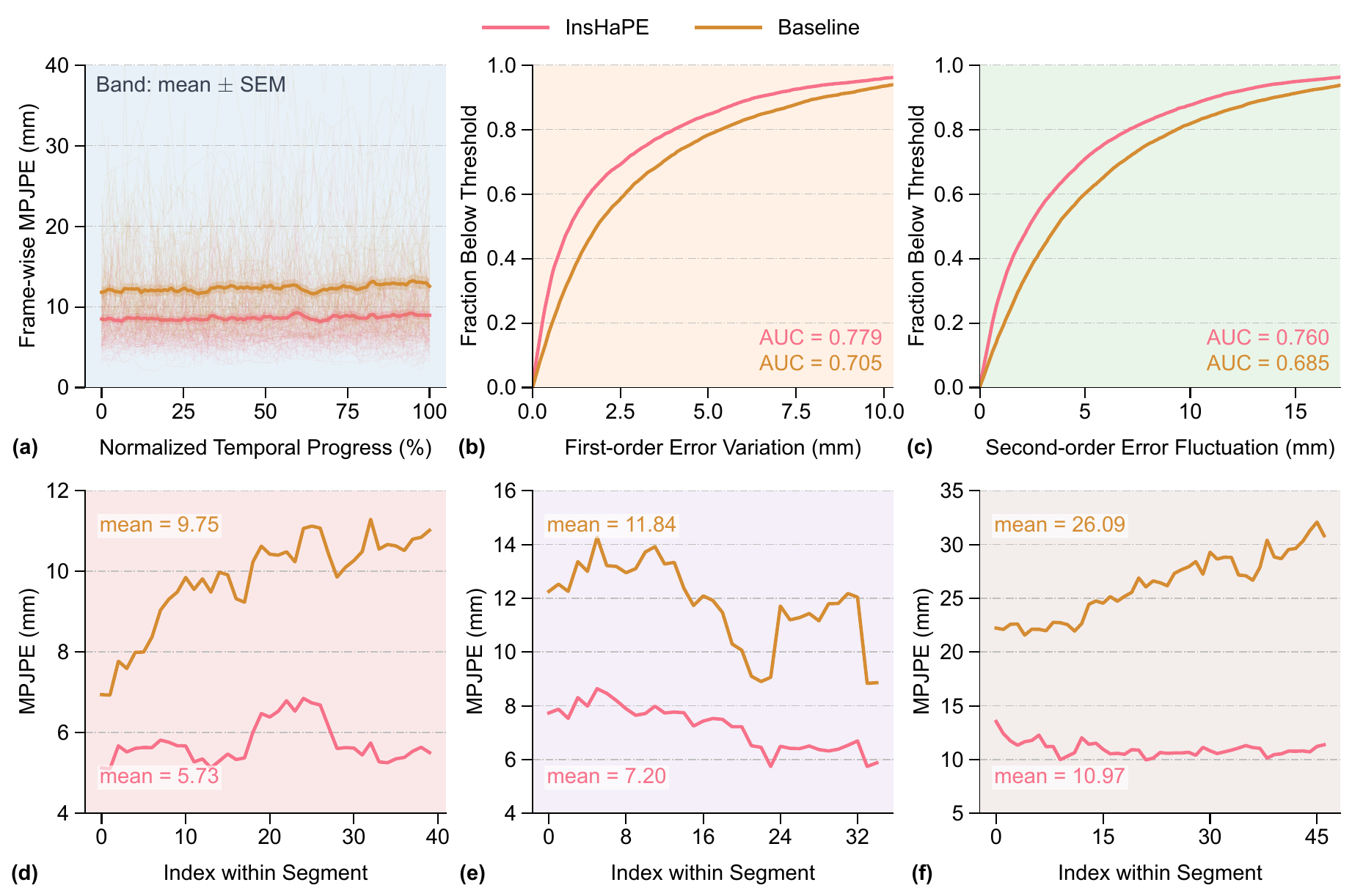}
    \MSfigcaption{fig:temporal_stability}{
    Temporal error variation analysis on IncNYU.
    (a) Frame-wise MPJPE across the inferred pseudo-continuous segments, with the mean trend and SEM.
    (b--c) Threshold-based cumulative distributions of the first-order error variation and second-order error fluctuation, respectively; curves closer to the upper-left indicate smaller variation of the scalar error.
    (d--f) Representative pseudo-continuous segments from low-, medium-, and high-error regimes, respectively.
    }{
    \MSphantomsubcaptions{%
    \phantomsubcaption\label{fig:temporal_stability-a}
    \phantomsubcaption\label{fig:temporal_stability-b}
    \phantomsubcaption\label{fig:temporal_stability-c}
    \phantomsubcaption\label{fig:temporal_stability-d}
    \phantomsubcaption\label{fig:temporal_stability-e}
    \phantomsubcaption\label{fig:temporal_stability-f}
    }
    }
\end{figure}

\myPara{Temporal Error Variation.}
To complement the frame-wise accuracy evaluation, we analyze the temporal behavior of the scalar pose error on IncNYU (see \cref{fig:temporal_stability}).
Since explicit sequence boundaries are unavailable in the NYU hand pose dataset \cite{tompson2014real}, we infer pseudo-continuous segments from the ground-truth motion: a boundary is placed between adjacent frames when the root displacement or the mean root-relative displacement of the 14 evaluation joints exceeds the median plus five times the median absolute deviation of that quantity over the test set, and all temporal comparisons are restricted to within these segments.
The missing-finger configuration of IncNYU varies across frames and is therefore not fixed within a segment; both methods are evaluated on the same frames and thus face the same configuration changes.
Specifically, 114 segments with at least 10 frames are retained, covering 7,383 of the 8,252 test frames (89.47\%).
As shown in \cref{fig:temporal_stability-a}, InsHaPE maintains consistently lower frame-wise MPJPE across the inferred segments.
The overall mean MPJPE decreases from 11.49 mm for the HandDAGT baseline \cite{cheng2024handdagt} to 8.00 mm for InsHaPE, corresponding to a relative reduction of 30.4\%.
We further characterize temporal error variation using the first-order difference
and the second-order difference,
which quantify adjacent-frame error variation and higher-order fluctuation of the error trajectory, respectively.
As shown in \cref{fig:temporal_stability-b,fig:temporal_stability-c}, the consistency curves of InsHaPE are consistently closer to the upper-left region.
The corresponding normalized AUC increases from 0.705 to 0.779 for the first-order measure and from 0.685 to 0.760 for the second-order measure, yielding relative improvements of 10.5\% and 11.0\%, respectively.
\cref{fig:temporal_stability-d,fig:temporal_stability-e,fig:temporal_stability-f} further present representative segments from low-, medium-, and high-error regimes.
Across all three regimes, InsHaPE exhibits both lower frame-wise errors and smaller variation of the scalar error than the baseline.
These results indicate that the scalar error of InsHaPE varies less over time. Since equal-magnitude offsets in opposite directions leave the scalar error unchanged, they do not measure the joint-level temporal stability.

\section{Prototype System and User Study}
\label{sec:system}

\begin{figure}
  \centering
  \includegraphics[width=\linewidth,clip,trim=70 120 70 120]{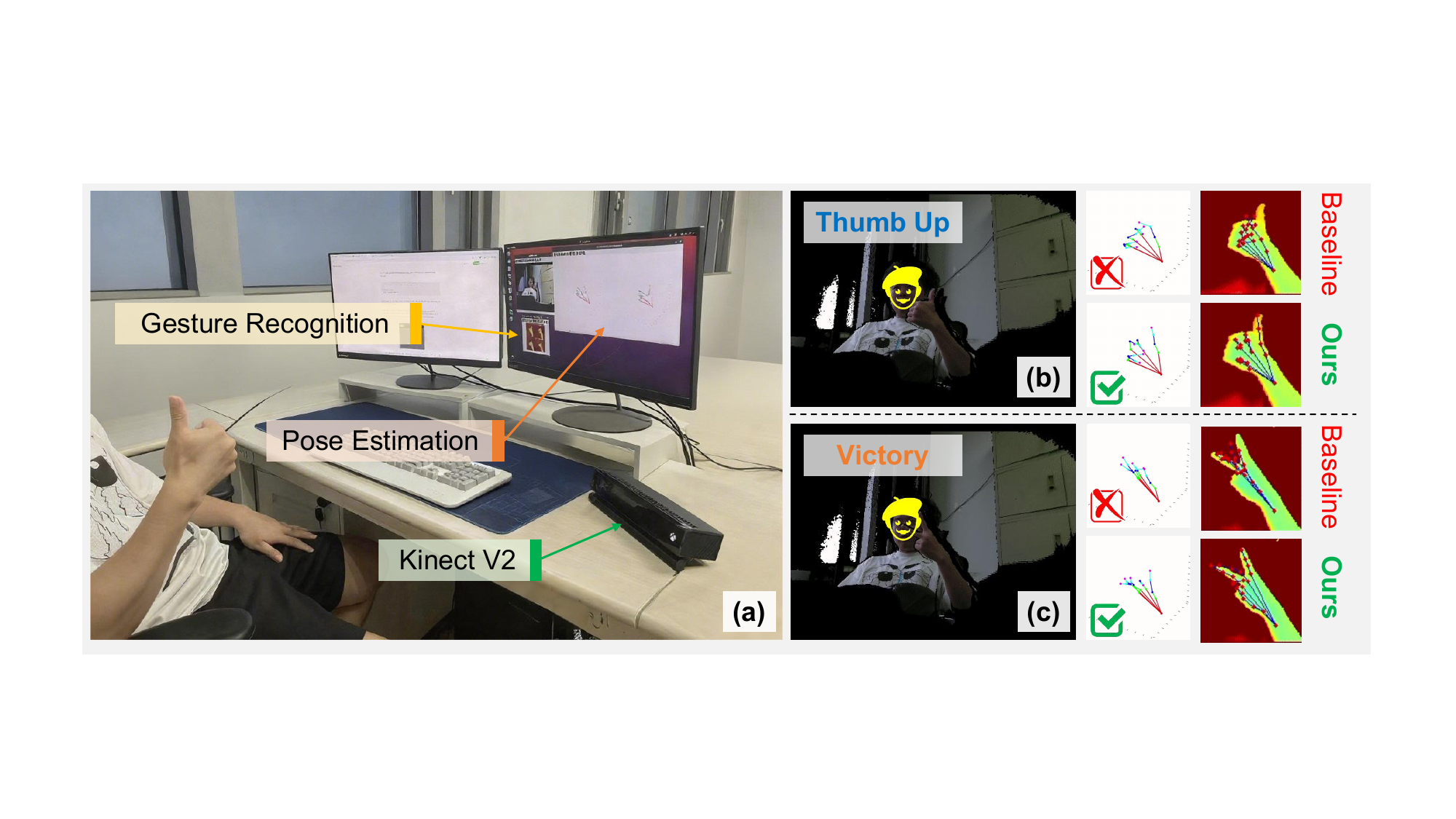}
  \caption{Prototype MR system powered by InsHaPE.
    (a) Hand pose estimation and interaction pipeline.
    (b--c) Representative interaction scenarios under structurally incomplete hand configurations.
    Please refer to the supplementary video for full demonstrations.}
  \label{fig:system}
  \MSphantomsubcaptions{%
    \phantomsubcaption\label{fig:system-a}
    \phantomsubcaption\label{fig:system-b}
    \phantomsubcaption\label{fig:system-c}
  }
\end{figure}

While \cref{sec:exp} demonstrates the accuracy and robustness of InsHaPE on benchmark datasets, it is important to examine whether these gains translate into practical benefits for downstream MR interaction.
Many MR systems use hand pose as an intermediate representation, where unstable or anatomically implausible skeletons can directly affect interaction quality and downstream motion understanding~\cite{zhou2021stgae}.
We therefore integrate InsHaPE into a prototype MR system (see \cref{fig:system}) and evaluate its downstream impact with the rule-based gesture recognizer of our previous MR system~\cite{zhou2021stgae}, which classifies each gesture from predefined distances between hand joints and is used without training or modification for both estimators.
We compare skeletons estimated by InsHaPE with those produced by the state-of-the-art baseline HandDAGT~\cite{cheng2024handdagt}, while keeping the downstream recognition framework unchanged.
We conduct two complementary evaluations: a controlled study with simulated persistent incompleteness and a pilot study involving users with real finger incompleteness.

\subsection{Controlled Study with Simulated Incompleteness}
\label{sec:simulated-user-study}

\begin{figure}
  \centering
  \includegraphics[width=\linewidth,clip,trim=50 180 50 180]{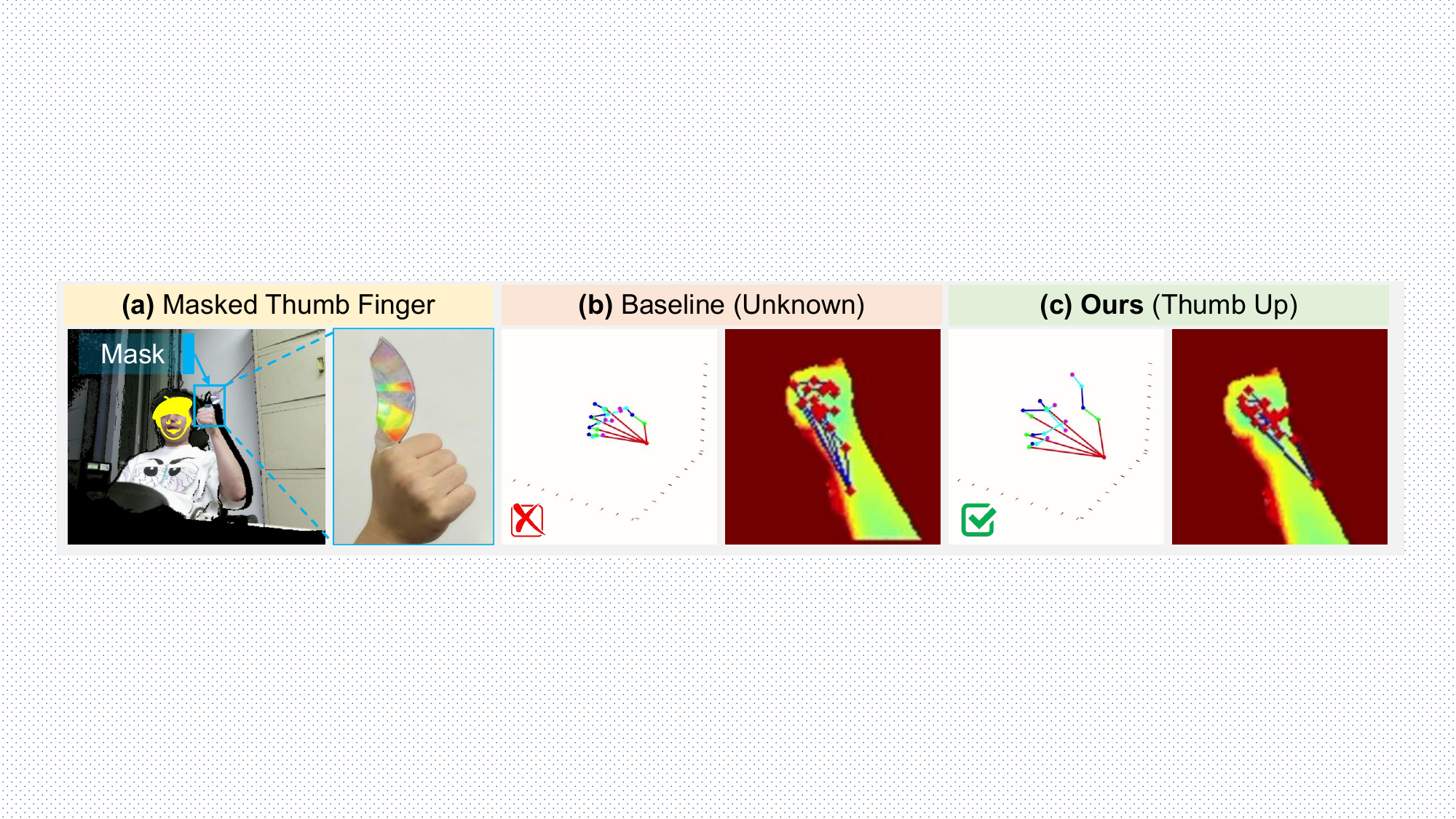}
  \caption{Controlled simulation of persistent missing observations using a light-sensitive finger mask.
    (a) Experimental setup.
    (b) HandDAGT~\cite{cheng2024handdagt} produces erroneous and implausible skeleton predictions under simulated incompleteness.
    (c) InsHaPE maintains more plausible and anatomically consistent skeleton estimation.}
  \label{fig:mask}
  \MSphantomsubcaptions{%
    \phantomsubcaption\label{fig:mask-a}
    \phantomsubcaption\label{fig:mask-b}
    \phantomsubcaption\label{fig:mask-c}
  }
\end{figure}

\myPara{Participants and Procedure.}
Eighteen university students (9 male, 9 female), aged 20--27 years (mean = 23.1, SD = 2.1), participated in the study.
All participants had normal or corrected-to-normal vision, reported no hand-related conditions, and provided informed consent. The study protocol received ethical approval before participant recruitment, and the approval documentation is included with the supplementary submission.
None had prior experience with the tested system, although most were familiar with MR devices.
To provide a controlled and reproducible setting, participants wore a lightweight physical mask on one finger to simulate persistent missing observations.
We emphasize that this setting simulates persistent visual absence on able-bodied participants and should not be interpreted as real anatomical incompleteness.
The study was conducted in a quiet laboratory environment (see \cref{fig:system-a}).
Participants first completed a short training session to familiarize themselves with the system.
The same protocol was used to evaluate InsHaPE and HandDAGT. Participants performed five predefined gestures (thumb up, pinch, open hand, fist, and victory) under two observation conditions:
(1) complete hand, and
(2) simulated incomplete hand with one finger masked (see \cref{fig:mask-a}).
The order of the two estimators and of the two observation conditions was counterbalanced across participants, and each gesture was repeated five times per condition.
Each estimator ran live in its own session, so the two methods processed separate performances of the same gestures rather than identical recordings.
Short breaks minimized fatigue; sessions lasted about 30 minutes.

\myPara{Evaluation Protocol.}
We evaluate the system from two perspectives.
(1) \emph{Objective performance}: gesture recognition accuracy, defined as the percentage of correctly classified gestures, is compared across pose estimation methods (InsHaPE vs.\ HandDAGT) and observation conditions (complete vs.\ simulated incomplete).
(2) \emph{Subjective user experience}: participants rated each estimator-based system separately on 7-point scales for perceived naturalness, ease of interaction, and reliability. Naturalness assesses consistency with hand motion, ease of interaction reflects effort, and reliability measures response consistency. Comparisons are paired by participant.

\begin{figure}
    \centering
    \includegraphics[width=\linewidth]{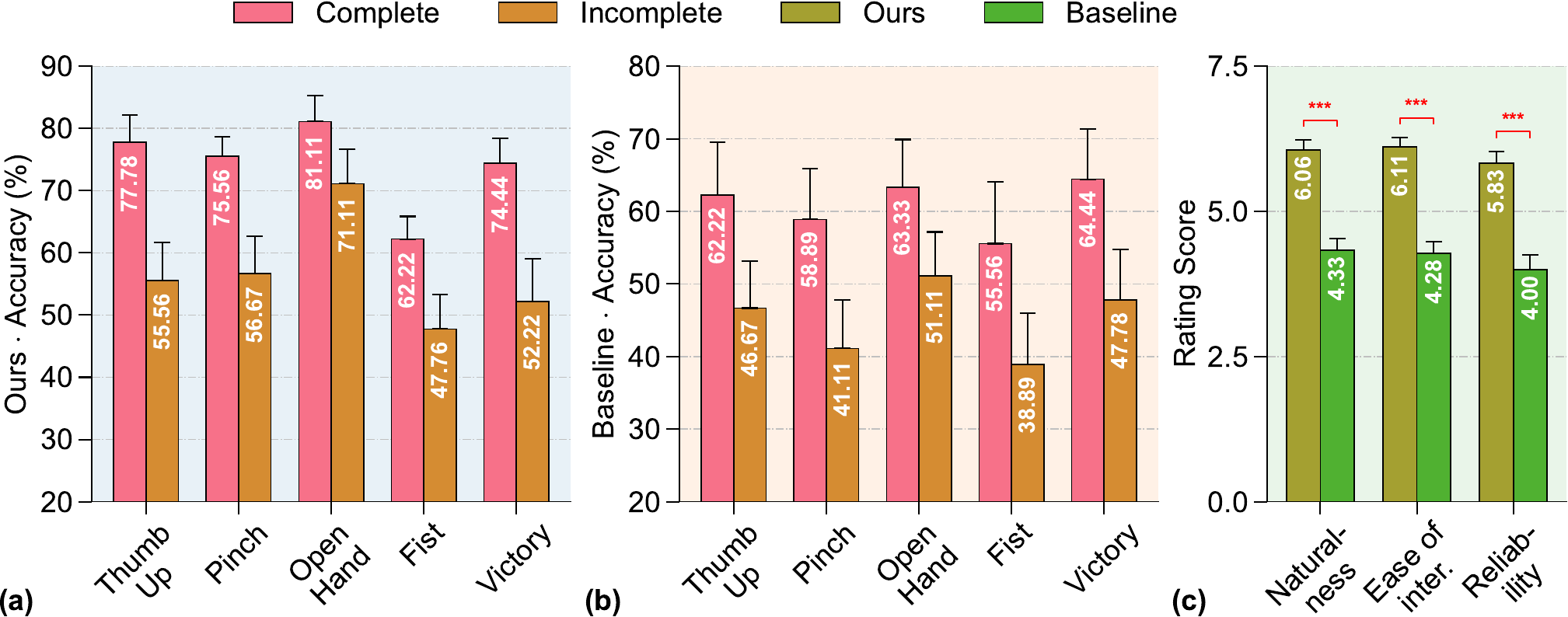}
    \MSfigcaption{fig:user_study}{Results of the controlled user study with simulated persistent incompleteness.
    (a--b) Gesture recognition accuracy under complete-hand and simulated incomplete-hand conditions for InsHaPE and HandDAGT, respectively.
    (c) Subjective ratings of perceived naturalness, ease of interaction, and reliability on a 7-point Likert scale.
    Bars show means, error bars denote standard errors of the means, and *** in (c) denotes $p<0.001$ under paired tests of the ratings.}{
        \MSphantomsubcaptions{%
            \phantomsubcaption\label{fig:user_study-a}
            \phantomsubcaption\label{fig:user_study-b}
            \phantomsubcaption\label{fig:user_study-c}
        }
    }
\end{figure}

\myPara{Results.}
Under complete-hand conditions, InsHaPE achieves higher recognition accuracy than HandDAGT (see \cref{fig:user_study-a,fig:user_study-b}), indicating that its structural modeling benefits downstream gesture recognition when the hand is fully observed. Under simulated incompleteness, HandDAGT suffers a substantial performance drop, whereas InsHaPE maintains higher recognition accuracy (see \cref{fig:user_study-a,fig:user_study-b}). Averaged over gestures, InsHaPE improves recognition accuracy by 13.3 and 11.6 percentage points under the complete and simulated incomplete settings, respectively. We report these differences descriptively as a system-level measure, without participant-level paired inference; since each estimator ran in its own live session, the differences also include variation between performances.
This suggests that more robust and anatomically consistent skeleton estimation can benefit downstream gesture recognition, particularly when local observations are persistently missing. Subjective ratings (mean $\pm$ SD) show a consistent advantage for InsHaPE across three dimensions (see \cref{fig:user_study-c}): naturalness (6.06\,$\pm$\,0.73 vs.\ 4.33\,$\pm$\,0.84), ease of interaction (6.11\,$\pm$\,0.68 vs.\ 4.28\,$\pm$\,0.83), and reliability (5.83\,$\pm$\,0.86 vs.\ 4.00\,$\pm$\,1.08). The paired mean differences, computed before rounding the displayed means, are 1.72 (95\% CI [1.35, 2.10]), 1.83 (95\% CI [1.48, 2.18]), and 1.83 (95\% CI [1.37, 2.29]), respectively, with all comparisons significant at $p<.001$ and large paired effect sizes ($d_z=2.29$, $2.59$, and $1.99$). Together, these results indicate that the robustness gains on IncNYU translate to more reliable downstream MR interaction under controlled conditions of persistent missing observations.

\setlength{\textfloatsep}{8pt plus 1pt minus 2pt}%
\begin{table}[t]
\centering
\setlength{\tabcolsep}{3pt}
\MStabcaption{tab:real_user}{
Pilot evaluation on three participants with real finger incompleteness.
We report gesture recognition accuracy (\%) and subjective ratings
on naturalness, ease of interaction, and reliability under the same downstream
MR protocol as the controlled study.
}
\resizebox{\linewidth}{!}{
\begin{tabular}{ccccccccc}
\toprule
\multirow{2.5}{*}{\textbf{Participant}}
& \multicolumn{2}{c}{\textbf{Recog. Acc.}}
& \multicolumn{2}{c}{\textbf{Naturalness}}
& \multicolumn{2}{c}{\textbf{Ease of Inter.}}
& \multicolumn{2}{c}{\textbf{Reliability}} \\
\cmidrule(lr){2-3}
\cmidrule(lr){4-5}
\cmidrule(lr){6-7}
\cmidrule(lr){8-9}
& Baseline & Ours
& Baseline & Ours
& Baseline & Ours
& Baseline & Ours \\
\midrule
P1
& 48.0 & \textbf{64.0}
& 5 & \textbf{6}
& 5 & \textbf{6}
& 4 & \textbf{6} \\

P2
& 44.0 & \textbf{60.0}
& 4 & \textbf{6}
& 4 & \textbf{6}
& 4 & \textbf{5} \\

P3
& 40.0 & \textbf{52.0}
& 4 & \textbf{5}
& 4 & \textbf{5}
& 3 & \textbf{5} \\
\midrule
Mean
& 44.0 & \textbf{58.7}
& 4.33 & \textbf{5.67}
& 4.33 & \textbf{5.67}
& 3.67 & \textbf{5.33} \\
\bottomrule
\end{tabular}
}%
\end{table}%

\MSsubsection{sec:real-user-study}{Real Incomplete-Hand Pilot Study}
\myPara{Participants and Procedure.}
To complement the controlled simulation with direct evidence of structural incompleteness, we recruited three male participants with finger incompleteness through the assistance of the Haidian District Disabled Persons' Federation in Beijing.
The participants were aged 28--35 years (mean = 32.0, SD = 3.6) and represented different forms of partial finger absence.
All participants signed written informed consent forms before the study, including consent for the collection and use of experimental data for research purposes.
We report this pilot separately from the controlled masked-user study because real anatomical absence differs fundamentally from simulated missing observations.
Each participant performed the predefined gestures using the MR pipeline and recognition protocol from \cref{sec:simulated-user-study}.

\begin{figure}
    \centering
    \includegraphics[width=\linewidth]{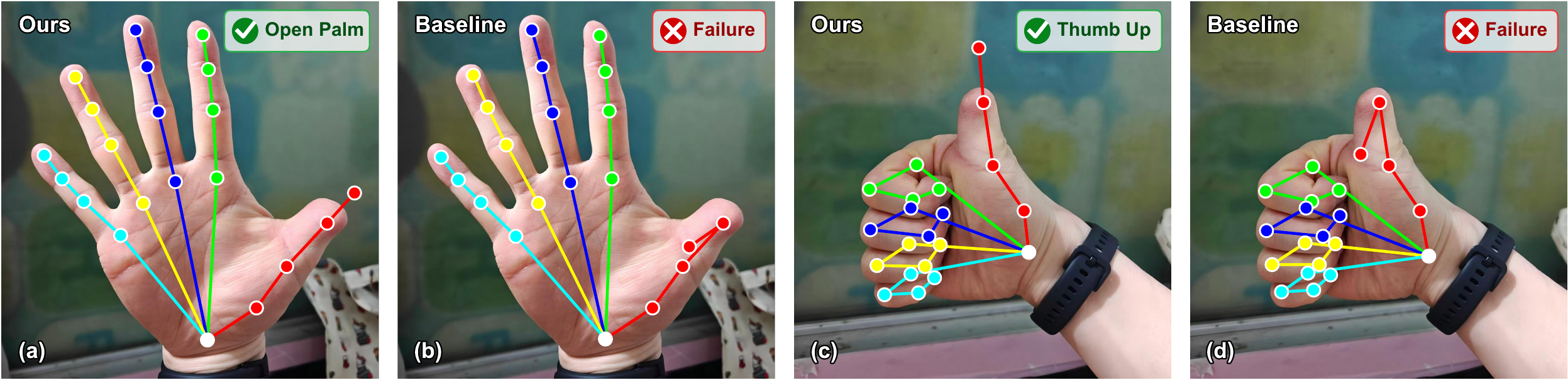}
    \MSfigcaption{fig:real_user_examples}{
    Representative results from the real incomplete-hand pilot study for a participant with partial thumb amputation.
    (a--b) Open-palm gesture estimated by InsHaPE and the HandDAGT baseline, respectively.
    (c--d) Thumb-up gesture estimated by InsHaPE and the HandDAGT baseline, respectively.
    In these trials, the downstream recognizer succeeds with InsHaPE and fails with the baseline.
    }{
    \MSphantomsubcaptions{%
      \phantomsubcaption\label{fig:real_user_examples-a}
      \phantomsubcaption\label{fig:real_user_examples-b}
      \phantomsubcaption\label{fig:real_user_examples-c}
      \phantomsubcaption\label{fig:real_user_examples-d}
    }
    }
\end{figure}

\myPara{Evaluation Protocol.}
We follow the same downstream evaluation protocol as in the controlled study to enable direct comparison.
Since absent joints in real incomplete hands do not have physical ground-truth locations, we do not compute phantom-joint errors.
Instead, we evaluate gesture recognition accuracy together with participant-reported naturalness, ease of interaction, and reliability on a 7-point Likert scale.
Given the small cohort, we report participant-wise results and descriptive averages without inferential statistical testing.

\myPara{Results.}
As shown in \cref{tab:real_user}, InsHaPE consistently improves downstream gesture recognition for all three participants, increasing the average accuracy from 44.0\% to 58.7\%.
The subjective ratings show the same trend: InsHaPE achieves higher average scores for naturalness (5.67 vs.\ 4.33), ease of interaction (5.67 vs.\ 4.33), and reliability (5.33 vs.\ 3.67).
Selected open-palm and thumb-up trials are visualized in \cref{fig:real_user_examples}, providing qualitative context for the participant-level results.
These consistent participant-level improvements provide preliminary evidence that the fixed-topology skeletons produced by InsHaPE remain useful for downstream MR interaction under genuine structural incompleteness.
Overall, the pilot complements the reproducible simulation with evidence from users with actual finger incompleteness.

\section{Conclusion and Discussion} \label{sec:conclusion}
This work revisits an assumption in hand pose estimation systems for MR: the availability of complete hand structures with a fixed topology. We present InsHaPE, an inclusive hand pose estimation framework that addresses structural incompleteness while maintaining performance for standard, fully articulated hands.
Extensive experiments on NYU/IncNYU and additional validation on MSRA demonstrate strong robustness to structural incompleteness, competitive performance on intact hands, and generalization beyond the primary benchmark.
These results show that explicitly modeling structural incompleteness improves the robustness of fixed-topology hand pose estimation to missing structures.
Beyond benchmark evaluation, a prototype MR system and user studies further demonstrate that InsHaPE provides more robust and topologically consistent skeletal representations for downstream interaction, improving gesture recognition performance and user-perceived interaction quality without modifying existing pipelines. This work emphasizes structural robustness in hand-centric perception and provides a practical step toward reliable MR interaction under incomplete hand configurations.

Despite these advances, the current approach still has several limitations. Topology-aware distillation relies on paired complete/incomplete observations during training; such pairs can be generated for IncNYU but are generally unavailable for users with real anatomical incompleteness. The present real-user pilot therefore provides downstream evidence only and does not remove this supervision dependency. This work also focuses on structural incompleteness and does not yet address other challenging factors such as motion instability or long-term fatigue. The model also has more parameters than lightweight baselines and lacks end-to-end evaluation on edge or wearable devices.
Although IncNYU provides a controlled and physiologically constrained benchmark, it is derived from NYU and cannot capture the full diversity of real hand anatomy, sensing conditions, and interaction behavior. Future work will investigate weakly supervised adaptation, broader hand variations, diverse real-world data, and efficient wearable deployment across practical scenarios.

\vspace{-40pt}

\begin{IEEEbiography}[{\includegraphics[width=1in,height=1.1in,clip,keepaspectratio]{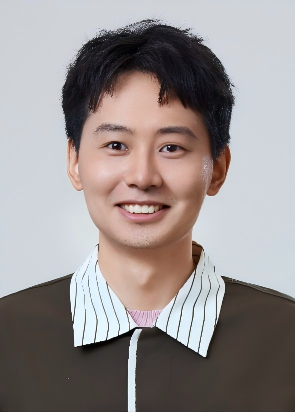}}]{Kanglei Zhou} 
is currently a Postdoctoral Researcher with the Department of Psychological and Cognitive Science, Tsinghua University. He received his Ph.D. degree in Computer Science and Engineering from Beihang University in 2025 and his B.E. degree from Henan Normal University in 2020. In 2024, he was a Visiting Student at Durham University. His research interests include human motion analysis and continual learning, with publications in leading venues such as TPAMI, IJCV, TIP, TVCG, ICML, ICLR, CVPR, ECCV, and ISMAR.
\end{IEEEbiography}

\vspace{-40pt}

\begin{IEEEbiography}[{\includegraphics[width=1in,height=1.1in,clip,trim=0 7 0 4,keepaspectratio]{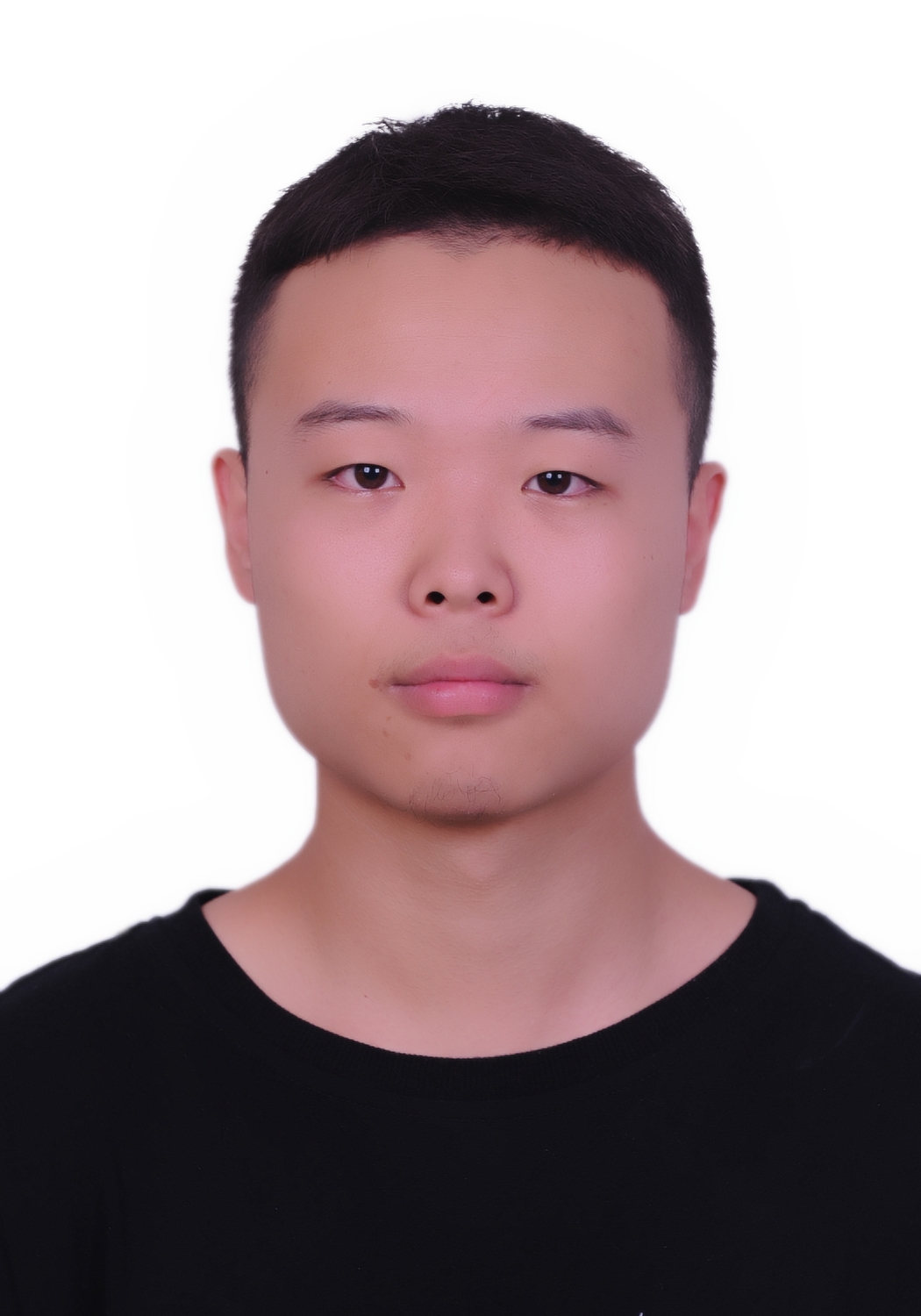}}]{Zikai Hao} received a B.Eng. degree from Taiyuan University of Technology. Now, he is pursuing a master of engineering in software engineering at the same school of Beihang University. He is currently engaged in human motion analysis and hand pose estimation.
\end{IEEEbiography}

\vspace{-40pt}

\begin{IEEEbiography}[{\includegraphics[width=1in,height=1.1in,clip,keepaspectratio]{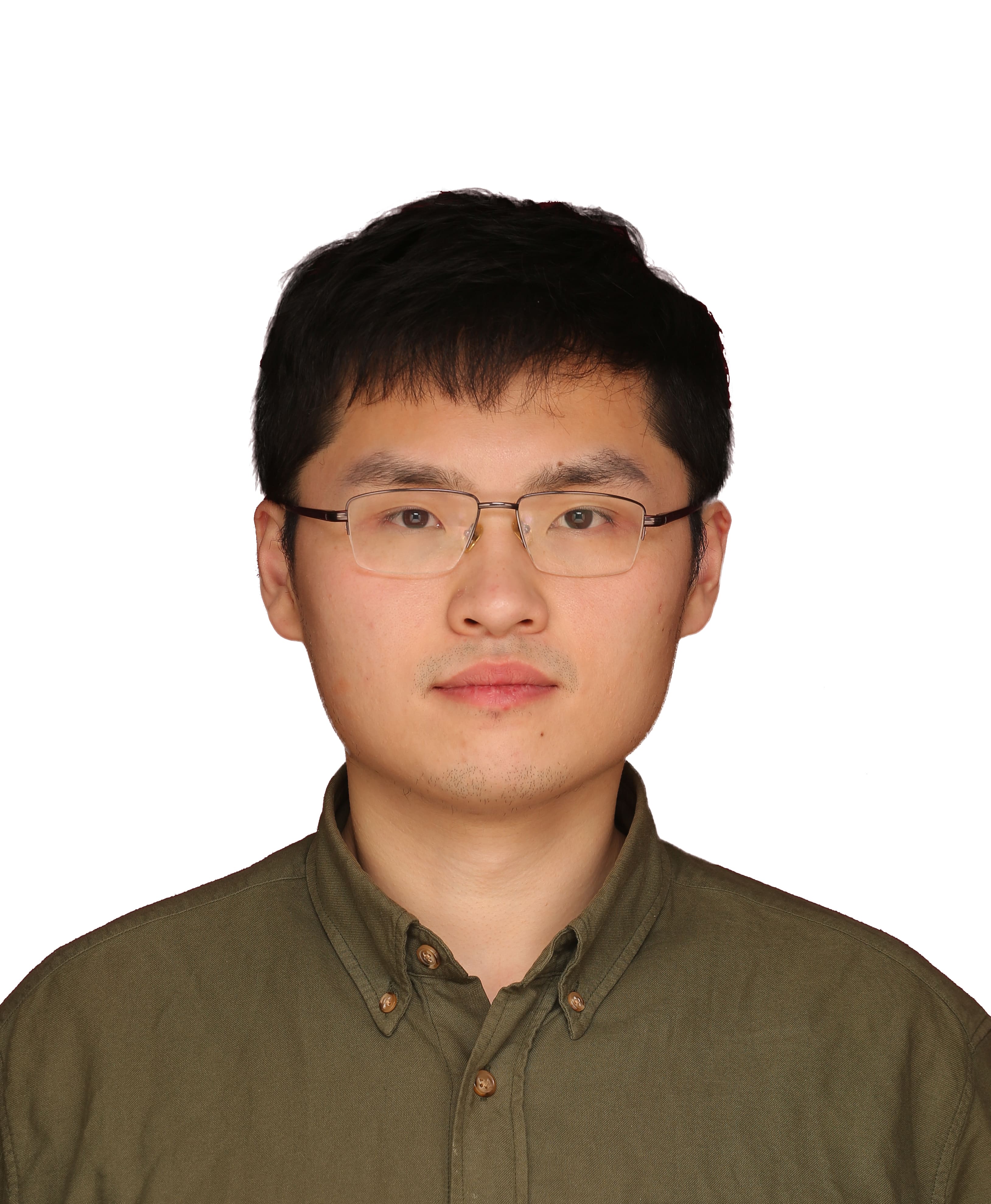}}]{Liyuan Wang}
    is currently an Assistant Professor in the Department of Psychological and Cognitive Sciences at Tsinghua University. He received the B.S. and Ph.D. degrees from Tsinghua University, where he also conducted his postdoctoral research. He has an interdisciplinary background in neuroscience and machine learning. His work on continual learning has been published in major conferences and journals in related fields, such as Nature Machine Intelligence, TPAMI, TNNLS, NeurIPS, ICLR, CVPR, ICCV, ECCV, etc.
\end{IEEEbiography}

\vspace{-40pt}

\begin{IEEEbiography}[{\includegraphics[width=1in,height=1.1in,clip,trim=85 260 85 80,keepaspectratio]{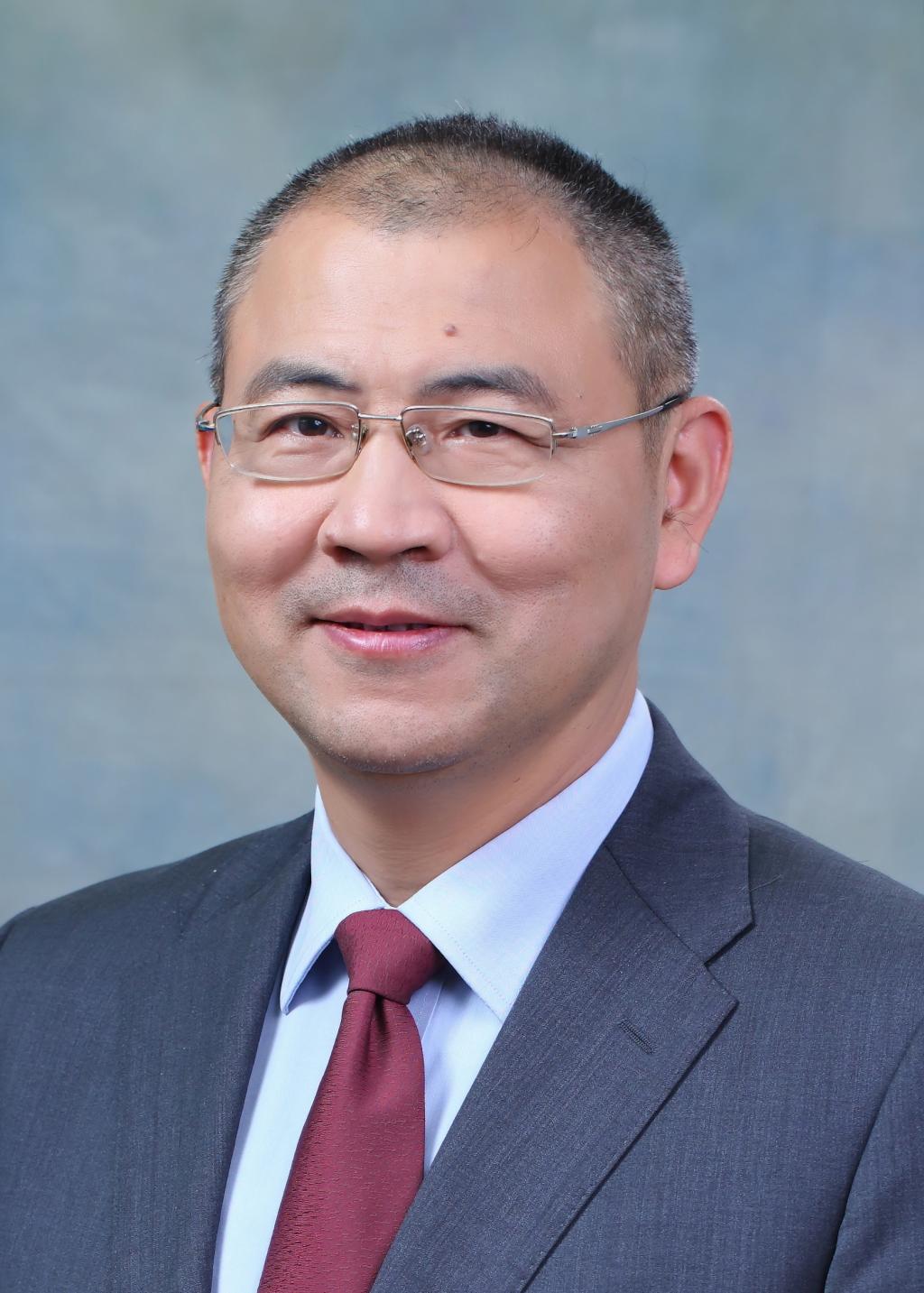}}]{Xiaohui Liang} received his Ph.D. degree in computer science and engineering from Beihang University, China. He is currently a Professor, working in the School of Computer Science and Engineering at Beihang University. His main research interests include computer graphics and animation, visualization, and virtual reality.
\end{IEEEbiography}


\begin{thebibliography}{10}

\bibitem{ohkawa2023efficient}
T.~Ohkawa, R.~Furuta, and Y.~Sato, ``Efficient annotation and learning for {3D}
  hand pose estimation: A survey,'' {\em International Journal of Computer
  Vision}, vol.~131, no.~12, pp.~3193--3206, 2023.

\bibitem{wan2018dense}
C.~Wan, T.~Probst, L.~Van~Gool, and A.~Yao, ``Dense {3D} regression for hand
  pose estimation,'' in {\em IEEE Conference on Computer Vision and Pattern
  Recognition}, pp.~5147--5156, 2018.

\bibitem{moon2022accurate}
G.~Moon, H.~Choi, and K.~M. Lee, ``Accurate {3D} hand pose estimation for
  whole-body {3D} human mesh estimation,'' in {\em IEEE/CVF Conference on
  Computer Vision and Pattern Recognition Workshops}, pp.~2308--2317, 2022.

\bibitem{salter2024emg2pose}
S.~Salter, R.~Warren, C.~Schlager, A.~Spurr, S.~Han, R.~Bhasin, Y.~Cai,
  P.~Walkington, A.~Bolarinwa, R.~J. Wang, {\em et~al.}, ``emg2pose: A large
  and diverse benchmark for surface electromyographic hand pose estimation,''
  {\em Advances in Neural Information Processing Systems}, vol.~37,
  pp.~55703--55728, 2024.

\bibitem{zhao2024gaussianhand}
L.~Zhao, X.~Lu, R.~Fan, S.~K. Im, and L.~Wang, ``{GaussianHand}: Real-time {3D}
  gaussian rendering for hand avatar animation,'' {\em IEEE Transactions on
  Visualization and Computer Graphics}, vol.~31, no.~9, pp.~6484--6496, 2025.

\bibitem{shi2022grasping}
Y.~Shi, L.~Zhao, X.~Lu, T.~Hoang, and M.~Wang, ``Grasping {3D} objects with
  virtual hand in {VR} environment,'' in {\em ACM SIGGRAPH International
  Conference on Virtual-Reality Continuum and Its Applications in Industry},
  pp.~1--8, 2022.

\bibitem{zhou2026comprehensive}
K.~Zhou, R.~Cai, L.~Wang, H.~P.~H. Shum, and X.~Liang, ``A comprehensive survey
  of action quality assessment: Method and benchmark,'' {\em Pattern
  Recognition}, vol.~179, p.~113933, 2026.

\bibitem{zhou2021stgae}
K.~Zhou, Z.~Cheng, H.~P. Shum, F.~W. Li, and X.~Liang, ``{STGAE}:
  Spatial-temporal graph auto-encoder for hand motion denoising,'' in {\em IEEE
  International Symposium on Mixed and Augmented Reality}, pp.~41--49, 2021.

\bibitem{cheng2024handdagt}
W.~Cheng, E.~Kim, and J.~H. Ko, ``{HandDAGT}: A denoising adaptive graph
  transformer for {3D} hand pose estimation,'' in {\em European Conference on
  Computer Vision}, vol.~15146, pp.~35--52, 2024.

\bibitem{ren2023two}
P.~Ren, Y.~Chen, J.~Hao, H.~Sun, Q.~Qi, J.~Wang, and J.~Liao, ``Two heads are
  better than one: Image-point cloud network for depth-based {3D} hand pose
  estimation,'' {\em AAAI Conference on Artificial Intelligence}, vol.~37,
  no.~2, pp.~2163--2171, 2023.

\bibitem{cheng2024handdiff}
W.~Cheng, H.~Tang, L.~Van~Gool, and J.~H. Ko, ``{HandDiff}: {3D} hand pose
  estimation with diffusion on image-point cloud,'' in {\em IEEE/CVF Conference
  on Computer Vision and Pattern Recognition}, pp.~2274--2284, 2024.

\bibitem{jiang2023a2j}
C.~Jiang, Y.~Xiao, C.~Wu, M.~Zhang, J.~Zheng, Z.~Cao, and J.~T. Zhou,
  ``{A2J-Transformer}: Anchor-to-joint transformer network for {3D} interacting
  hand pose estimation from a single {RGB} image,'' in {\em IEEE/CVF Conference
  on Computer Vision and Pattern Recognition}, pp.~8846--8855, 2023.

\bibitem{mo2021gesture}
G.~B. Mo, J.~J. Dudley, and P.~O. Kristensson, ``Gesture knitter: A hand
  gesture design tool for head-mounted mixed reality applications,'' in {\em
  CHI Conference on Human Factors in Computing Systems}, pp.~1--13, 2021.

\bibitem{zhao2022place}
L.~Zhao, X.~Lu, Q.~Bao, and M.~Wang, ``In-place gestures classification via
  long-term memory augmented network,'' in {\em IEEE International Symposium on
  Mixed and Augmented Reality}, pp.~224--233, 2022.

\bibitem{zhou2023mixed}
K.~Zhou, C.~Chen, Y.~Ma, Z.~Leng, H.~P. Shum, F.~W. Li, and X.~Liang, ``A mixed
  reality training system for hand-object interaction in simulated microgravity
  environments,'' in {\em IEEE International Symposium on Mixed and Augmented
  Reality}, pp.~167--176, 2023.

\bibitem{ohkawa2023assemblyhands}
T.~Ohkawa, K.~He, F.~Sener, T.~Hodan, L.~Tran, and C.~Keskin,
  ``{AssemblyHands}: Towards egocentric activity understanding via {3D} hand
  pose estimation,'' in {\em IEEE/CVF Conference on Computer Vision and Pattern
  Recognition}, pp.~12999--13008, 2023.

\bibitem{li2023renderih}
L.~Li, L.~Tian, X.~Zhang, Q.~Wang, B.~Zhang, L.~Bo, M.~Liu, and C.~Chen,
  ``{RenderIH}: A large-scale synthetic dataset for {3D} interacting hand pose
  estimation,'' in {\em IEEE/CVF International Conference on Computer Vision},
  pp.~20395--20405, 2023.

\bibitem{zhou2024multi}
K.~Zhou, H.~P. Shum, F.~W. Li, and X.~Liang, ``Multi-task spatial-temporal
  graph auto-encoder for hand motion denoising,'' {\em IEEE Transactions on
  Visualization and Computer Graphics}, vol.~30, no.~10, pp.~6754--6769, 2024.

\bibitem{ye2018occlusion}
Q.~Ye and T.-K. Kim, ``Occlusion-aware hand pose estimation using hierarchical
  mixture density network,'' in {\em European Conference on Computer Vision},
  vol.~11214, pp.~817--834, 2018.

\bibitem{meng20223d}
H.~Meng, S.~Jin, W.~Liu, C.~Qian, M.~Lin, W.~Ouyang, and P.~Luo, ``{3D}
  interacting hand pose estimation by hand de-occlusion and removal,'' in {\em
  European Conference on Computer Vision}, vol.~13666, pp.~380--397, 2022.

\bibitem{liu2022explicit}
Q.~Liu, Y.~Zhang, S.~Bai, and A.~Yuille, ``Explicit occlusion reasoning for
  multi-person {3D} human pose estimation,'' in {\em European Conference on
  Computer Vision}, vol.~13665, pp.~497--517, 2022.

\bibitem{park2022handoccnet}
J.~Park, Y.~Oh, G.~Moon, H.~Choi, and K.~M. Lee, ``{HandOccNet}:
  Occlusion-robust {3D} hand mesh estimation network,'' in {\em IEEE/CVF
  Conference on Computer Vision and Pattern Recognition}, pp.~1496--1505, 2022.

\bibitem{sun2015cascaded}
X.~Sun, Y.~Wei, S.~Liang, X.~Tang, and J.~Sun, ``Cascaded hand pose
  regression,'' in {\em IEEE Conference on Computer Vision and Pattern
  Recognition}, pp.~824--832, 2015.

\bibitem{deng2022recurrent}
X.~Deng, D.~Zuo, Y.~Zhang, Z.~Cui, J.~Cheng, P.~Tan, L.~Chang, M.~Pollefeys,
  S.~Fanello, and H.~Wang, ``Recurrent {3D} hand pose estimation using cascaded
  pose-guided {3D} alignments,'' {\em IEEE Transactions on Pattern Analysis and
  Machine Intelligence}, vol.~45, no.~1, pp.~932--945, 2023.

\bibitem{leng2021stable}
Z.~Leng, J.~Chen, H.~P. Shum, F.~W. Li, and X.~Liang, ``Stable hand pose
  estimation under tremor via graph neural network,'' in {\em IEEE Virtual
  Reality and 3D User Interfaces}, pp.~226--234, 2021.

\bibitem{tompson2014real}
J.~Tompson, M.~Stein, Y.~LeCun, and K.~Perlin, ``Real-time continuous pose
  recovery of human hands using convolutional networks,'' {\em ACM Transactions
  on Graphics}, vol.~33, no.~5, pp.~1--10, 2014.

\bibitem{du2019crossinfonet}
K.~Du, X.~Lin, Y.~Sun, and X.~Ma, ``{CrossInfoNet}: Multi-task information
  sharing based hand pose estimation,'' in {\em IEEE/CVF Conference on Computer
  Vision and Pattern Recognition}, pp.~9896--9905, 2019.

\bibitem{fang2020jgr}
L.~Fang, X.~Liu, L.~Liu, H.~Xu, and W.~Kang, ``{JGR-P2O}: Joint graph reasoning
  based pixel-to-offset prediction network for {3D} hand pose estimation from a
  single depth image,'' in {\em European Conference on Computer Vision},
  vol.~12351, pp.~120--137, 2020.

\bibitem{ge2016robust}
L.~Ge, H.~Liang, J.~Yuan, and D.~Thalmann, ``Robust {3D} hand pose estimation
  in single depth images: from single-view {CNN} to multi-view {CNNs},'' in
  {\em IEEE Conference on Computer Vision and Pattern Recognition},
  pp.~3593--3601, 2016.

\bibitem{ren2021pose}
P.~Ren, H.~Sun, J.~Hao, Q.~Qi, J.~Wang, and J.~Liao, ``Pose-guided hierarchical
  graph reasoning for {3-D} hand pose estimation from a single depth image,''
  {\em IEEE Transactions on Cybernetics}, vol.~53, no.~1, pp.~315--328, 2023.

\bibitem{ge20173d}
L.~Ge, H.~Liang, J.~Yuan, and D.~Thalmann, ``{3D} convolutional neural networks
  for efficient and robust hand pose estimation from single depth images,'' in
  {\em IEEE Conference on Computer Vision and Pattern Recognition},
  pp.~1991--2000, 2017.

\bibitem{moon2018v2v}
G.~Moon, J.~Y. Chang, and K.~M. Lee, ``{V2V-PoseNet}: Voxel-to-voxel prediction
  network for accurate {3D} hand and human pose estimation from a single depth
  map,'' in {\em IEEE Conference on Computer Vision and Pattern Recognition},
  pp.~5079--5088, 2018.

\bibitem{qi2017pointnet}
C.~R. Qi, H.~Su, K.~Mo, and L.~J. Guibas, ``{PointNet}: Deep learning on point
  sets for {3D} classification and segmentation,'' in {\em IEEE Conference on
  Computer Vision and Pattern Recognition}, pp.~652--660, 2017.

\bibitem{ge2018hand}
L.~Ge, Y.~Cai, J.~Weng, and J.~Yuan, ``Hand {PointNet}: {3D} hand pose
  estimation using point sets,'' in {\em IEEE Conference on Computer Vision and
  Pattern Recognition}, pp.~8417--8426, 2018.

\bibitem{lin2023cross}
Q.~Lin, L.~Yang, and A.~Yao, ``Cross-domain {3D} hand pose estimation with dual
  modalities,'' in {\em IEEE/CVF Conference on Computer Vision and Pattern
  Recognition}, pp.~17184--17193, 2023.

\bibitem{ge2025coarse}
Y.~Ge, C.~Xu, and L.~Cheng, ``A coarse-to-fine multi-hypothesis method for
  ambiguous hand pose estimation,'' {\em IEEE Transactions on Image
  Processing}, vol.~34, pp.~4302--4314, 2025.

\bibitem{ho2020denoising}
J.~Ho, A.~Jain, and P.~Abbeel, ``Denoising diffusion probabilistic models,''
  {\em Advances in Neural Information Processing Systems}, vol.~33,
  pp.~6840--6851, 2020.

\bibitem{sohl2015deep}
J.~Sohl-Dickstein, E.~Weiss, N.~Maheswaranathan, and S.~Ganguli, ``Deep
  unsupervised learning using nonequilibrium thermodynamics,'' in {\em
  International Conference on Machine Learning}, vol.~37 of {\em Proceedings of
  Machine Learning Research}, pp.~2256--2265, 2015.

\bibitem{ivashechkin2023denoising}
M.~Ivashechkin, O.~Mendez, and R.~Bowden, ``Denoising diffusion for {3D} hand
  pose estimation from images,'' in {\em IEEE/CVF International Conference on
  Computer Vision Workshops}, pp.~3136--3145, 2023.

\bibitem{zhang2023ochid}
S.~Zhang, T.~Zheng, Z.~Chen, J.~Hu, A.~Khamis, J.~Liu, and J.~Luo,
  ``{OCHID-Fi}: Occlusion-robust hand pose estimation in {3D} via
  {RF}-vision,'' in {\em IEEE/CVF International Conference on Computer Vision},
  pp.~15112--15121, 2023.

\bibitem{garg2025occrobnet}
M.~Garg, P.~M. Pradhan, and D.~Ghosh, ``{OccRobNet}: Occlusion robust network
  for accurate {3D} interacting hand-object pose estimation,'' in {\em National
  Conference on Communications}, pp.~1--6, 2025.

\bibitem{yang2025occlusion}
H.~Yang, W.~Sun, J.~Liu, J.~Zheng, J.~Xiao, and A.~Mian, ``Occlusion-aware {3D}
  hand-object pose estimation with masked autoencoders,'' {\em IEEE
  Transactions on Multimedia}, pp.~1--11, 2026.

\bibitem{cai2025geometry}
Y.~Cai, H.~Pan, J.~Yang, Y.~Liu, Q.~Gao, and X.~Wang, ``Geometry-aware {3D}
  hand--object pose estimation under occlusion via hierarchical feature
  decoupling,'' {\em Electronics}, vol.~14, no.~5, p.~1029, 2025.

\bibitem{woo2023survey}
T.~Woo, W.~Park, W.~Jeong, and J.~Park, ``A survey of deep learning methods and
  datasets for hand pose estimation from hand-object interaction images,'' {\em
  Computers \& Graphics}, vol.~116, pp.~474--490, 2023.

\bibitem{dudley2023inclusive}
J.~Dudley, L.~Yin, V.~Garaj, and P.~O. Kristensson, ``Inclusive immersion: a
  review of efforts to improve accessibility in virtual reality, augmented
  reality and the metaverse,'' {\em Virtual Reality}, vol.~27, no.~4,
  pp.~2989--3020, 2023.

\bibitem{inclusiveAgenda2023}
C.~Creed, M.~Al-Kalbani, A.~Theil, S.~Sarcar, and I.~Williams, ``Inclusive
  augmented and virtual reality: A research agenda,'' {\em International
  Journal of Human--Computer Interaction}, vol.~40, no.~20, pp.~6200--6219,
  2024.

\bibitem{creed2024inclusive}
C.~Creed, M.~Al-Kalbani, A.~Theil, S.~Sarcar, and I.~Williams, ``Inclusive
  {AR/VR}: accessibility barriers for immersive technologies,'' {\em Universal
  Access in the Information Society}, vol.~23, no.~1, pp.~59--73, 2024.

\bibitem{biswas2021adaptive}
P.~Biswas, P.~Orero, M.~Swaminathan, K.~Krishnaswamy, and P.~Robinson,
  ``Adaptive accessible {AR/VR} systems,'' in {\em CHI Conference on Human
  Factors in Computing Systems}, pp.~1--7, 2021.

\bibitem{krug2024don}
K.~Krug, J.~M{\'e}ndez, W.~Luo, and R.~Dachselt, ``Don't leave me out:
  Designing for device inclusivity in mixed reality collaboration,'' {\em arXiv
  Preprint arXiv:2409.05374}, 2024.

\bibitem{kim2025immersive}
J.~Kim, J.-H. Ahn, and Y.~Kim, ``Immersive interaction for inclusive virtual
  reality navigation: Enhancing accessibility for socially underprivileged
  users,'' {\em Electronics}, vol.~14, no.~5, p.~1046, 2025.

\bibitem{othman2024accessible}
A.~Othman, K.~Chemnad, A.~E. Hassanien, A.~Tlili, C.~Y. Zhang, D.~Al-Thani,
  F.~Alt{\i}nay, H.~Chalghoumi, H.~S. Al-Khalifa, M.~Obeid, {\em et~al.},
  ``Accessible metaverse: A theoretical framework for accessibility and
  inclusion in the metaverse,'' {\em Multimodal Technologies and Interaction},
  vol.~8, no.~3, p.~21, 2024.

\bibitem{radanliev2024accessibility}
P.~Radanliev, D.~De~Roure, P.~Novitzky, and I.~Sluganovic, ``Accessibility and
  inclusiveness of new information and communication technologies for disabled
  users and content creators in the metaverse,'' {\em Disability and
  Rehabilitation: Assistive Technology}, vol.~19, no.~5, pp.~1849--1863, 2024.

\bibitem{cheng2023handr2n2}
W.~Cheng and J.~H. Ko, ``{HandR2N2}: Iterative {3D} hand pose estimation using
  a residual recurrent neural network,'' in {\em IEEE/CVF International
  Conference on Computer Vision}, pp.~20904--20913, 2023.

\bibitem{choi2022learning}
H.~Choi, G.~Moon, J.~Park, and K.~M. Lee, ``Learning to estimate robust {3D}
  human mesh from in-the-wild crowded scenes,'' in {\em IEEE/CVF Conference on
  Computer Vision and Pattern Recognition}, pp.~1475--1484, 2022.

\bibitem{everingham2011pascal}
M.~Everingham and J.~Winn, ``The {PASCAL} visual object classes challenge 2011
  ({VOC2011}) development kit.'' Technical report, 2011.

\bibitem{rezaei2023trihorn}
M.~Rezaei, R.~Rastgoo, and V.~Athitsos, ``{TriHorn-Net}: a model for accurate
  depth-based {3D} hand pose estimation,'' {\em Expert Systems with
  Applications}, vol.~223, p.~119922, 2023.

\bibitem{kawaiah2020fingertip}
A.~Kawaiah, M.~Thakur, S.~Garg, S.~H. Kawasmi, and A.~Hassan, ``Fingertip
  injuries and amputations: a review of the literature,'' {\em Cureus},
  vol.~12, no.~5, p.~e8291, 2020.

\bibitem{armagan2020measuring}
A.~Armagan, G.~Garcia-Hernando, S.~Baek, S.~Hampali, M.~Rad, Z.~Zhang, S.~Xie,
  M.~Chen, B.~Zhang, F.~Xiong, {\em et~al.}, ``Measuring generalisation to
  unseen viewpoints, articulations, shapes and objects for {3D} hand pose
  estimation under hand-object interaction,'' in {\em European Conference on
  Computer Vision}, vol.~12368, pp.~85--101, 2020.

\bibitem{park2019relational}
W.~Park, D.~Kim, Y.~Lu, and M.~Cho, ``Relational knowledge distillation,'' in
  {\em IEEE/CVF Conference on Computer Vision and Pattern Recognition},
  pp.~3967--3976, 2019.

\bibitem{guo2017towards}
H.~Guo, G.~Wang, X.~Chen, C.~Zhang, F.~Qiao, and H.~Yang, ``Region ensemble
  network: Towards good practices for deep {3D} hand pose estimation,'' {\em
  Journal of Visual Communication and Image Representation}, vol.~55,
  pp.~404--414, 2018.

\bibitem{oberweger2017deepprior++}
M.~Oberweger and V.~Lepetit, ``{DeepPrior++}: Improving fast and accurate {3D}
  hand pose estimation,'' in {\em IEEE International Conference on Computer
  Vision Workshops}, pp.~585--594, 2017.

\bibitem{chen2020pose}
X.~Chen, G.~Wang, H.~Guo, and C.~Zhang, ``Pose guided structured region
  ensemble network for cascaded hand pose estimation,'' {\em Neurocomputing},
  vol.~395, pp.~138--149, 2020.

\bibitem{xiong2019a2j}
F.~Xiong, B.~Zhang, Y.~Xiao, Z.~Cao, T.~Yu, J.~T. Zhou, and J.~Yuan, ``{A2J}:
  Anchor-to-joint regression network for {3D} articulated pose estimation from
  a single depth image,'' in {\em IEEE/CVF International Conference on Computer
  Vision}, pp.~793--802, 2019.

\bibitem{huang2020awr}
W.~Huang, P.~Ren, J.~Wang, Q.~Qi, and H.~Sun, ``{AWR}: Adaptive weighting
  regression for {3D} hand pose estimation,'' {\em AAAI Conference on
  Artificial Intelligence}, vol.~34, no.~07, pp.~11061--11068, 2020.

\bibitem{ren2021spatial}
P.~Ren, H.~Sun, W.~Huang, J.~Hao, D.~Cheng, Q.~Qi, J.~Wang, and J.~Liao,
  ``Spatial-aware stacked regression network for real-time {3D} hand pose
  estimation,'' {\em Neurocomputing}, vol.~437, pp.~42--57, 2021.

\bibitem{huang2020hand}
L.~Huang, J.~Tan, J.~Liu, and J.~Yuan, ``{Hand-Transformer}: Non-autoregressive
  structured modeling for {3D} hand pose estimation,'' in {\em European
  Conference on Computer Vision}, vol.~12370, pp.~17--33, 2020.

\bibitem{ge2018point}
L.~Ge, Z.~Ren, and J.~Yuan, ``Point-to-point regression {PointNet} for {3D}
  hand pose estimation,'' in {\em European Conference on Computer Vision},
  vol.~11217, pp.~489--505, 2018.

\bibitem{cheng2021handfoldingnet}
W.~Cheng, J.~H. Park, and J.~H. Ko, ``{HandFoldingNet}: A {3D} hand pose
  estimation network using multiscale-feature guided folding of a {2D} hand
  skeleton,'' in {\em IEEE/CVF International Conference on Computer Vision},
  pp.~11260--11269, 2021.

\bibitem{li2018visualizing}
H.~Li, Z.~Xu, G.~Taylor, C.~Studer, and T.~Goldstein, ``Visualizing the loss
  landscape of neural nets,'' {\em Advances in Neural Information Processing
  Systems}, vol.~31, pp.~6391--6401, 2018.

\end{thebibliography}
\end{document}